\documentclass[journal]{IEEEtran}

\usepackage{mathrsfs}
\usepackage{amsmath,amsfonts,amsthm,amssymb}

\theoremstyle{plain} \newtheorem{definition}{Definition}
\theoremstyle{plain}  \newtheorem{theorem}{Theorem}

\usepackage{booktabs}
\usepackage{algorithmicx}
\usepackage{algpseudocode}
\usepackage[citecolor=blue, colorlinks]{hyperref}
\usepackage{svg}
\usepackage[numbers,sort&compress]{natbib}
\usepackage{subfigure}

\usepackage[ruled,vlined]{algorithm2e}
\usepackage{array}
\usepackage{textcomp}
\usepackage{stfloats}
\usepackage{url}
\usepackage{verbatim}
\usepackage{graphicx}
\usepackage{cite}
\newcommand{\revice}[1]{\textcolor{black}{#1}}
\usepackage{xcolor}
\usepackage{soul}
\usepackage{tikz}
\usepackage{arydshln}
\usepackage{newtxtext}

\title{Irregular Tensor Low-Rank Representation for Hyperspectral Image Representation}

\author{Bo Han,~Yuheng Jia,~\IEEEmembership{Member,~IEEE},~Hui Liu, and Junhui Hou,~\IEEEmembership{Senior Member,~IEEE}
\thanks{This work was supported part by the National Natural Science Foundation of China under Grants U24A20322 and 62422118, and part by Hong Kong UGC under grants UGC/FDS11/E02/22 and UGC/FDS11/E03/24. This research work was also supported by the Big Data Computing Center of Southeast University.  \textit{(Corresponding author: Yuheng Jia.)}

Bo Han is with the School of Computer Science and Engineering, Southeast University, Nanjing 210096, China (e-mail: hanbo@seu.edu.cn). 

Yuheng Jia is with the School of Computer Science and Engineering, Southeast University, Nanjing 210096, China, and also with the Key Laboratory of New Generation Artificial Intelligence Technology and Its Interdisciplinary Applications (Southeast University), Ministry of Education, China, and the School of Computing Information Sciences, Saint Francis University, Hong Kong (e-mail: yhjia@seu.edu.cn). 

Hui Liu is with the School of Computing Information Sciences, Saint Francis University, Hong Kong (e-mail: hliu99-c@my.cityu.edu.hk). 

Junhui Hou is with the Department of Computer Science, City University of Hong Kong, Hong Kong (e-mail: jh.hou@cityu.edu.hk).}
\thanks{}}

{}

\begin{document}

\maketitle

\begin{abstract}
Spectral variations pose a common challenge in analyzing hyperspectral images (HSI). To address this, low-rank tensor representation has emerged as a robust strategy, leveraging inherent correlations within HSI data. However, the spatial distribution of ground objects in HSIs is inherently irregular, existing naturally in tensor format, with numerous class-specific regions manifesting as irregular tensors. Current low-rank representation techniques are designed for regular tensor structures and overlook this fundamental irregularity in real-world HSIs, leading to performance limitations. To tackle this issue, we propose a novel model for irregular tensor low-rank representation tailored to efficiently model irregular 3D cubes. {By incorporating a non-convex nuclear norm to promote low-rankness and integrating a global negative low-rank term to enhance the discriminative ability, our proposed model is formulated as a constrained optimization problem and solved using an alternating augmented Lagrangian method.} Experimental validation conducted on four public datasets demonstrates the superior performance of our method compared to existing state-of-the-art approaches. The code is publicly available at \url{https://github.com/hb-studying/ITLRR}.

\end{abstract}


\section{Introduction}

\begin{figure}[!t]
\centering
\subfigure[]{\includegraphics[width=0.9in]{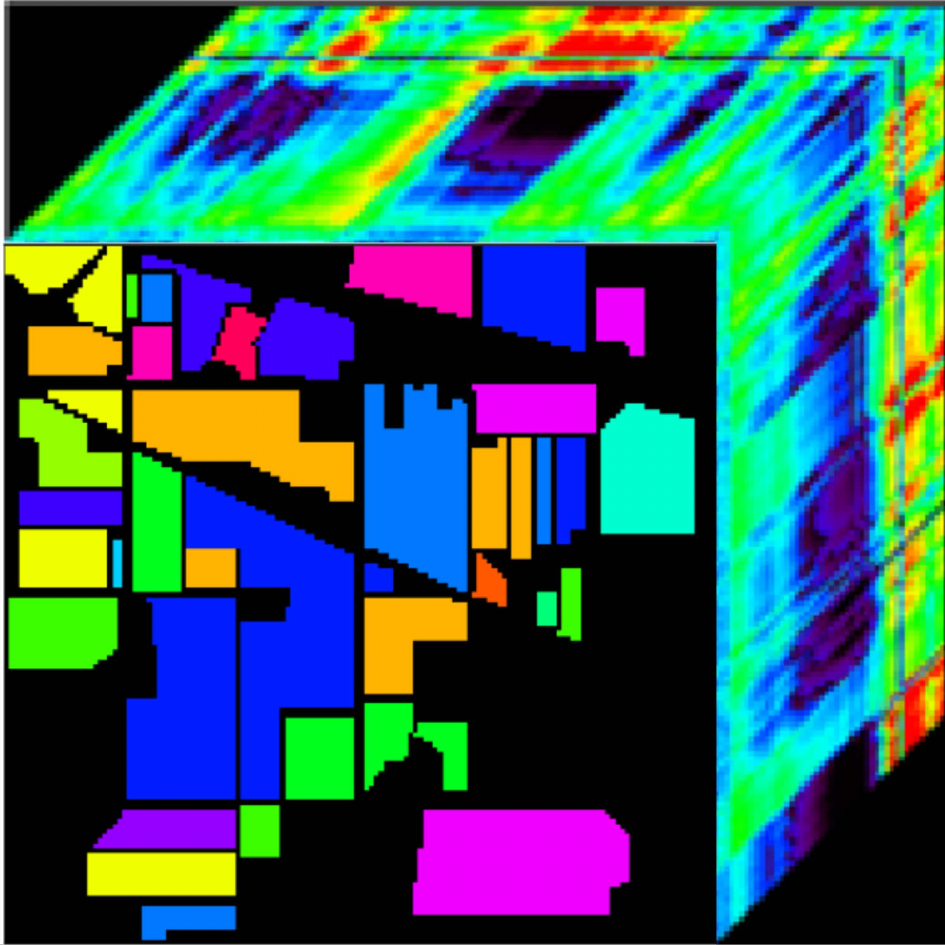}\label{subfig_grand-truth}}%
\hfil
\subfigure[]{\includegraphics[width=0.9in]{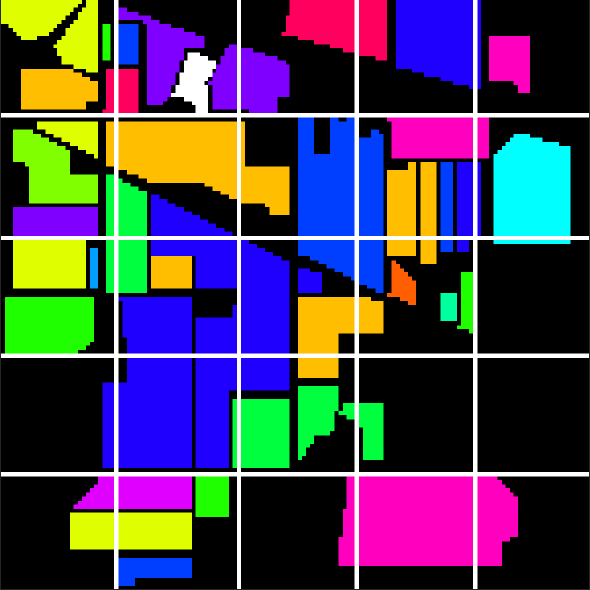}\label{subfig_patched}}%
\hfil
\subfigure[]{\includegraphics[width=0.9in]{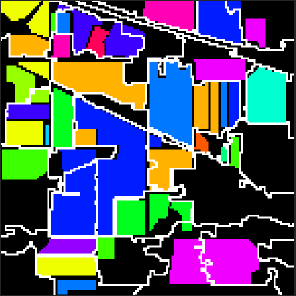}\label{subfig_ers}}%
\caption{(a) presents a 3D representation of hyperspectral data, showcasing the distribution of ground truth. (b) displays an image patched into blocks. (c) illustrates an image segmented into superpixels, allowing for a more effective representation of spatial structures.}
\label{fig_section1}
\end{figure}

 With the advancement of hyperspectral remote sensing technology, hyperspectral images (HSIs) can capture spectral data across hundreds of contiguous bands, as shown in Fig. \ref{subfig_grand-truth}.  This rich band information enables discrimination among various materials and facilitates applications in agriculture \citep{agricultural}, environmental monitoring \citep{environment_monitor2} and urban planning \citep{urban_planning}. Since HSIs typically contain a limited set of materials and exhibit high correlation among their spectral signatures, they inherently possess a low-rank structure \citep{structure_of_HSI}. However, due to sensor interference or variations in imaging conditions, the spectral signatures of pixels in homogeneous regions may vary considerably \citep{2014HRSI}, disrupting this low-rank structure.

To address spectral variation and enhance representation ability, numerous methods leveraging the low-rank property of HSIs have been proposed.  Several methods \citep{2016HRSI-RPCA,2018FRPCALG} are based on Robust Principal Component Analysis (RPCA \citep{2011rpca}) that unfold three-dimensional hyperspectral images into matrices, assuming that the matrix can be decomposed into a low-rank matrix and a sparse noise matrix. Later, considering that similar materials usually appear in local regions, as illustrated in Fig. \ref{subfig_grand-truth}, many methods \citep{2014ILRA,2015-adjust-LRMR,2017SSLR,2018S3LRR,GLRB} partition the HSI into rectangular patches as illustrated in Fig. \ref{subfig_patched} and restore each patch individually using RPCA-based techniques. To capture local information more effectively, superpixel segmentation algorithms have been employed to obtain irregular homogeneous regions in \citep{2017SSLRR,2018SLRA_GBM,2017Multiscale_SLRA,SP-DLRR} as illustrated in Fig. \ref{subfig_ers}.

Nonetheless, it is essential to recognize that the data structure of HSI is intrinsically tensor-based, not matrix-based \citep{2014HRSI}. The above methods transform HSI data into matrix form, potentially corrupting the spatial features \citep{2020TRPCA}. In contrast, tensor representations can preserve spatial information effectively and yield superior restoration results. Tensor-based robust principal component analysis (TRPCA) models decompose the HSI tensor into low-rank and sparse components, which are widely applied to HSI \citep{trpca_13,trpca_14,trpca_15,trpca_11,2020LSSTRPCA,2022LPGTRPCA}. However, it is important to note that the above tensor-based methods can only handle regular data cubes. While as shown in Fig. \ref{subfig_grand-truth}, the same material in HSI is typically distributed in irregular, localized regions, the global tensor low-rank representation applied to the entire HSI cube is not well-suited for capturing the irregular data distribution.

\begin{figure*}
    \centering
    \includegraphics[width=1\textwidth]{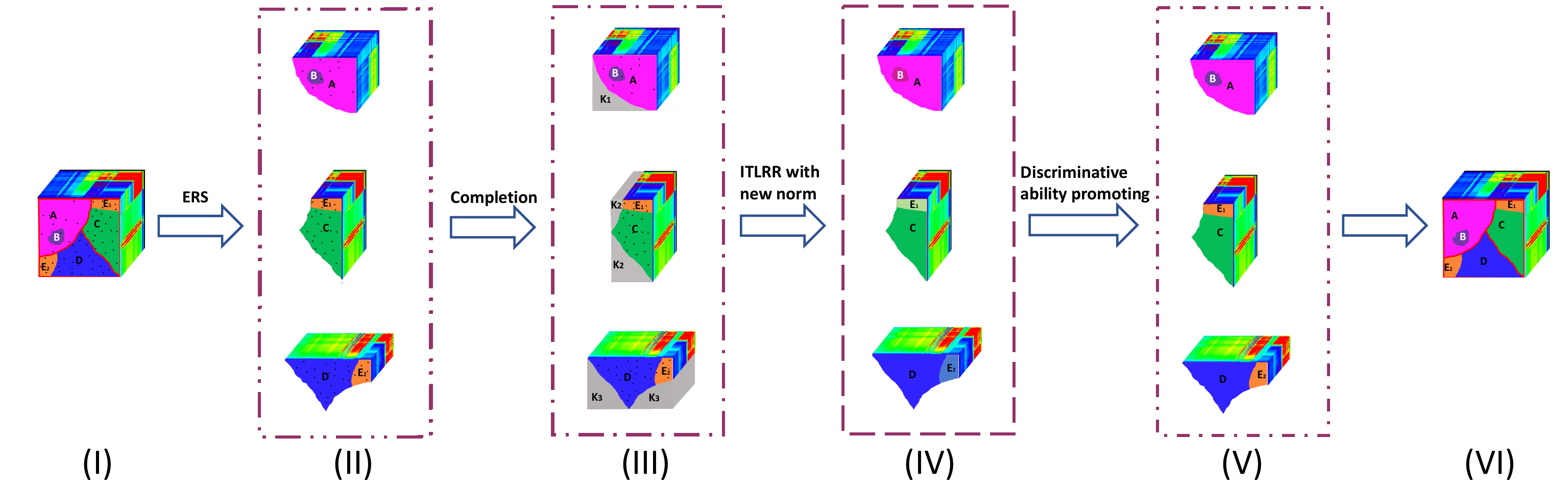}
    \caption{Framework of the proposed method. (I) stands for the origin HSI with spectral variation. The same color indicates that those pixels belong to the same class, and the black dot represents the data noise. We first divide the input HSI into several irregular 3D cubes by a typical superpixel segmentation method ERS; Then we complete the irregular 3D data cubes with black shape-complementary data cubes to constitute regular tensor patches (III), and design a novel ITLRR model (IV) that can pursue the tensor low-rank representation, which only relies on the original irregular data cubes and ignores the black complementary 3D patches. Besides, the ITLRR is based on a non-convex tensor nuclear norm that can better approximate the tensor low-rankness (IV). The above irregular tensor low-rank representation processes the 3D patches individually, which may lead to the over-smoothing problem, i.e., two disconnected areas with the same material do not share a similar representation (e.g., E1 and E2 in (IV)). To address the problem, we propose to add a negative low-rank term on the whole HSI to enhance the overall discriminative ability (V). Finally, we pack the low-ranked irregular 3D cubes to get the final low-rank HSI representation (VI).}
    \label{mind}
\end{figure*}

To this end, as shown in Fig. \ref{mind}, we propose a novel Irregular Tensor Low-Rank Representation (ITLRR), which can effectively process the irregular three-dimensional data cubs. For the irregular 3D data, we first complete it by using blank tensors to make up a regular data cube, and then we apply distinct low-rank and sparse constraints to the  complementary data cubes and the original irregular data cube, which enables the effective exploration of the low-rank representation of irregular 3D data cube. We further design a non-convex tensor nuclear norm that closely approximates the true rank, and propose a global negative low-rank term to enhance discriminative ability of the overall low-rank representation. The proposed model is formulated as a optimization problem and solved by alternative augmented Lagrangian method. Extensive experimental results demonstrate that our method surpasses state-of-the-art low-rank based and deep-learning based methods significantly. The main contributions of this study are summarized as follows:

\begin{enumerate}
\item{Our proposed method is the first to explore the discriminative low-rank properties of the irregular tensors.}
\item{We introduce a non-convex nuclear norm to pursue low-rank representation of the local irregular 3D cubes and introduce a negative low-rank term to avoid the over-smoothing representation issue.}

\end{enumerate}

The rest of this paper is organized as follows. Section \ref{sec:notation and related_works} introduces the notations and preliminaries used in this paper, along with a review of existing methods for HSI. The proposed ITLRR method is presented in Section \ref{sec:proposed-method}. Section \ref{sec:experiments} provides experimental evaluations and comparisons, followed by the conclusions in Section \ref{sec:conclusion}.

\begin{table}[ht]
\caption{Notations}
\label{NOTATIONS}
\begin{tabular}{l}
\toprule
 \\[-6pt]

1. $\overline{\mathcal{A}} $ is the discrete Fast Fourier Transform (FFT) of $\mathcal{A}$, i.e.,\\ [5pt]\quad $\overline{\mathcal{A}} = fft(\mathcal{A}, [], 3)$ and $\mathcal{A} = ifft(\overline{\mathcal{A}}, [], 3)$ where $fft$ denotes the FFT \\ [5pt]\quad  operator and $ifft$ denotes the inverse FFT operator.\\ [5pt] 

2. $\overline{\mathbf{A}}^{(i)} \in \mathbb{R}^{n_1 \times n_2 }$ is the $i$-th frontal slice of $\overline{\mathcal{A}}\in \mathbb{R}^{n_1 \times n_2 \times n_3}$.\\ [5pt] 


3. $\mathbf{A}_{(3)} \in \mathbb{R}^{n_1 n_2 \times n_3}$ is the third order unfolding of $\mathcal{A} \in \mathbb{R}^{n_1 \times n_2 \times n_3}$,\\ [5pt]\quad which can be expressed as $\mathscr{F}_3\left(\mathbf{A}_{(3)}\right) = \mathcal{A}$.\\[5pt]

4. $\langle\mathcal{A}, \mathcal{B}\rangle:=\sum_i \sum_j \sum_k a_{i j k} b_{i j k}$. \\[5pt]

5. $\|\mathcal{A}\|_F:=\sqrt{\sum_i \sum_j \sum_k a_{i j k}^2}$. \\[5pt]

6. $\|\mathcal{A}\|_1:=\sum_i \sum_j \sum_k\left|a_{i j k}\right|$. \\[5pt]

7. $\|\mathcal{A}\|_{\star}:=\|\mathbf{A}_{(3)}\|_*$ with $\|\cdot\|_*$ being the nuclear norm of the 2D matrix.\\[5pt]

\bottomrule
\end{tabular}
\end{table}

\section{Preliminary and Related Work }\label{sec:notation and related_works}

HSIs are widely used in various applications due to their rich spectral information. However, due to imperfect imaging conditions, HSIs often suffer from the spectral variation problem, where the same material may exhibit different spectral signatures. Low-rank representation is an effective approach to address this issue. In this section, we first introduce the notations and preliminaries related to tensor low-rank representation, followed by a review of existing methods for HSIs.

\subsection{Notations and Preliminaries}

In this paper, we denote scalars by lowercase letters, e.g., $a$, vectors by bold lowercase letters, e.g., $\mathbf{a}$, 2-D matrices by bold uppercase letters, e.g., $\mathbf{A}$, and 3D tensors by calligraphic uppercase letters, e.g., $\mathcal{A}$.  And $a_{i j k}$ is the $(i, j, k)_{th}$ entry of $\mathcal{A}$ and  $\mathbf{A}^{(k)}$ is the $k$-th frontal slice of $\mathcal{A}$. More notations are summarized in Table \ref{NOTATIONS}. We will introduce some preliminaries for the tensor nuclear norm.

\begin{theorem}
Let \( \mathbf{A} \in  \mathbb{R}^{ n_1 \times n_2 } \) be a matrix, {the Singular Value Decomposition (SVD)} of \( \mathbf{A} \) is a factorization of the form:
\begin{equation}
 \mathbf{A} = \mathbf{U} \mathbf{S} \mathbf{V}^T,  
\end{equation}
where \( \mathbf{U} \in  \mathbb{R}^{ n_1 \times n_1 } \) and \( \mathbf{V}\in  \mathbb{R}^{ n_2 \times n_2 } \) are orthogonal matrices, \( \mathbf{S}  \in  \mathbb{R}^{ n_1 \times n_2 }\) is a diagonal matrix.
\end{theorem}

\begin{theorem}\citep{theorem3}
    Let $\mathbf{A} \in \mathbb{R}^{n_1 \times n_2}$ be an arbitrary matrix. The partial derivative of $\|\mathbf{A}\|_{*}$ is $\partial\|\mathbf{A}\|_{*} = \{\mathbf{UV^T} + \mathbf{Z} :  \mathbf{Z} \in \mathbb{R}^{n_1 \times n_2}, \mathbf{U^TZ} = 0, \mathbf{ZV} = 0, \|\mathbf{Z}\|_2 \leq 1\}$, where $\mathbf{A} =  \mathbf{USV^T}$ and $\| \mathbf{Z}\|_2 = \sqrt{\sum^{n_1}_i \sum^{n_2}_j  a_{i j }^2} $.
    \label{linearize}
\end{theorem}

\begin{theorem}\citep{theorem4}
     For any $\lambda>0$ and $\mathbf{Y} \in \mathbb{R}^{n_1 \times n_2}$, a globally optimal solution to the following optimization problem
\begin{equation}
\min_{\mathbf{X}} \lambda \sum_{i=1}^{s} f(\sigma_{i}(\mathbf{X}))+\frac{1}{2}\|\mathbf{X}-\mathbf{Y}\|_{F}^{2},
\end{equation}
is given by the weighted singular value thresholding:
\begin{equation}
\mathbf{X}^{*}=\mathbf{U} \mathbf{S}_{\epsilon} \mathbf{V}^\mathbf{T},
\end{equation}
where $s = \min \left(n_{1}, n_{2}\right)$, $f(\cdot): \mathbb{R} \rightarrow \mathbb{R}^{+}$ is continuous, concave and monotonically increasing on $[0, \infty)$ and $\sigma_{i}(\cdot)$ denotes the $i$-th largest singular value of a matrix. $\mathbf{Y}=\mathbf{U} \mathbf{S} \mathbf{V}^\mathbf{T}$,  $\mathbf{S}_{\epsilon}=$ $\operatorname{Diag}\left\{\left(\mathbf{S}_{i i}-\lambda \partial f(\sigma_{i}(\mathbf{X}))\right)_{+}\right\}$ and $k_+$ denotes the positive part of $k$, i.e, $k_+ = max(k,0)$. 
\label{optial}
\end{theorem}

\begin{definition}[\textbf{Tensor nuclear norm}]
For a tensor $\mathcal{A} \in \mathbb{R}^{n_1 \times n_2 \times n_3}$, its nuclear norm defined in \citep{2020TRPCA} is
\begin{equation}
\|\mathcal{A}\|_{*} = \sum_{i=1}^{n 3}\|\overline{\mathbf{A}}^{(i)}\|_{*} = \sum_{i=1}^{n 3} \sum_{j=1}^{l} \sigma_{j}(\overline{\mathbf{A}}^{(i)}),
\end{equation}
where $l = \min \left(n_{1}, n_{2}\right)$.
\label{definition_nuclear_norm}
\end{definition}

\begin{definition}[\textbf{The $\oplus$ operator}]
{For a regular tensor $\mathcal{X} \in \mathbb{R}^{n_1 \times n_2 \times n_3}$, composed of several non-overlapping irregular tensors $\mathcal{X}_1, \mathcal{X}_2, \dots, \mathcal{X}_k$, it can be expressed as $\mathcal{X} = \mathcal{X}_1 \oplus \mathcal{X}_2 \oplus \cdots \oplus \mathcal{X}_k$. Specially, the regular tensor is indexed by a matrix $\mathcal{I} \in \mathbb{R}^{n_1 \times n_2}$. Each irregular tensor $\mathcal{X}_i$ is associated with a unique index set $\mathcal{I}_i$, i.e, $\mathcal{I}_i \cap \mathcal{I}_j = \emptyset, \forall i \neq j$ and the union of all index sets satisfies $\bigcup_{i=1}^k \mathcal{I}_i = \mathcal{I}$. Each index set $\mathcal{I}_i$ determines the positions of the $i$-th irregular tensor $\mathcal{X}_i$ in the regular tensor $\mathcal{X}$, i.e, $\mathcal{X}(a, b, :) = \mathcal{X}_i(a, b, :),  \forall (a, b) \in \mathcal{I}_i$.
The $\oplus$ operator is then defined to merge these irregular tensors based on their index mappings:}
\begin{equation}
\mathcal{X}_i \oplus \mathcal{X}_j = \mathcal{X}(a, b, :),  \forall (a, b) \in  \mathcal{I}_i \cup \mathcal{I}_j.
\end{equation}

\label{definition_tensor_merging}
\end{definition}

\begin{figure}[ht]
    \centering
    \includegraphics[width=0.95\linewidth]{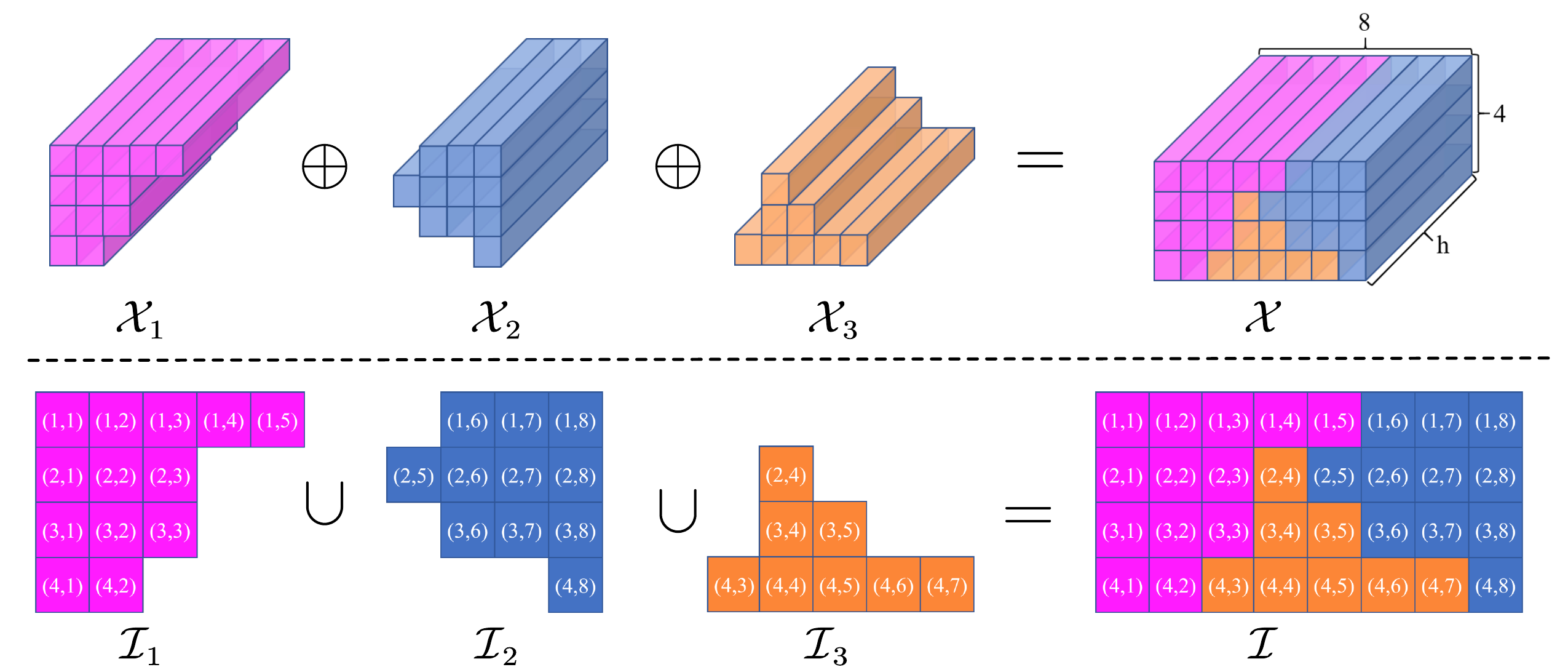}
    \caption{The illustration of the operation $\oplus$ to combine three irregular 3D patches.}
    \label{fig_oper_define_r3}
\end{figure}

{For example, for a regular tensor $\mathcal{X} \in \mathbb{R}^{8 \times 4 \times h}$ composed of three irregular tensors, the process of merging based on the $\oplus$ operation is illustrated in Fig. \ref{fig_oper_define_r3}.}

\subsection{Related Work}

Due to the inherent low-rank property of HSIs, many low-rank-based methods have been proposed to remedy the spectral variation problem. For instance, \citet{2016HRSI-RPCA} restructured HSI data into a two-dimensional matrix and applied RPCA to capture the low-rank representation, effectively reducing spectral variations. Additionally, \citet{2018FRPCALG} assumed that HSIs exhibit a clustering property within a unified subspace and incorporated a Laplacian graph term into the RPCA model to enhance the low-rank matrix generation.

The methods mentioned above do not reveal the local spatial correlation of pixel points within the same region. To address this, several approaches have been proposed to leverage this spatial prior. For example, \citet{2014ILRA} divided the HSIs into patches and restored each patch separately using the Go Decomposition algorithm \citep{2011GoDec}. Considering the variability of noise intensity across different bands, \citet{2015-adjust-LRMR} introduced an iterative framework with noise adjustments employing a patch-wise low-rank approximation method for HSI. \citet{2017SSLR} employed RPCA on patches to explore local spatial information and applied RPCA individually on each spectral band to investigate spatial information. Moreover, \citet{2018S3LRR} proposed a model exploring the low-rank property spectral and spatial domain simultaneously. Since homogeneous regions are often irregular, \citet{2017SSLRR} and \citet{2018SLRA_GBM} incorporated superpixel segmentation to preserve the spatial correlation of pixels in homogeneous regions. \citet{2017Multiscale_SLRA} proposed a strategy of multiscale superpixels to solve the problem of determining the optimal superpixel size.

{The above matrix-based methods may corrupt spatial features due to the inherent 3D structure of HSIs. To address this, many tensor-based methods have been proposed. \citet{tubal_rank}, and \citet{2014TRPCA} designed the tensor multi-rank and the tensor tubal rank respectively. \citet{2020TRPCA} defined a new tensor nuclear norm consistent with the matrix nuclear norm and proposed a TRPCA model that decomposes the 3D HSI tensor into a low-rank tensor and an error tensor. \citet{trpca_15} considered both the outliers and different types of noise and proposed an HSI denoising model based on the robust low-rank tensor recovery. \citet{trpca_11} incorporated spectral graph regularization into TRPCA and proposed the graph-regularized TRPCA algorithm for HSI denoising. \citet{2020LSSTRPCA} proposed a lateral-slice sparse tensor RPCA with a tensor $l_{2,1}$ norm to model sparse components and handle gross errors or outliers. \citet{2022LPGTRPCA} incorporated a tensor $l_{2,2,1}$ norm for frontal slice sparsity and a position-based Laplacian graph to preserve the local structure, improving hyperspectral image classification performance. \citet{Compare_review} proposed a tensor convolution-like low-rank dictionary framework that accounts for the shift-invariant low-rankness of tensor data and integrated it into a TRPCA model. \citet{revivwer-2} and \citet{revivwer-2-1} considered the irregularity of the real data distribution; however, they overlooked the discriminability of the representation. }


 While tensor SVD-based methods have shown promising results in capturing the low-rank structure of HSIs, other tensor decomposition techniques, such as Tucker decomposition and Tensor Ring (TR) decomposition, have also gained significant attention in recent years \citep{tensor_decom_review}. For instance, \citet{tucker_superreolution} introduced a hyperspectral image super-resolution framework based on Tucker decomposition, incorporating $l_1$-norm regularization on the core tensor to model sparsity and unidirectional total variation on dictionary matrices to ensure piecewise smoothness. Similarly, \citet{tucker_denoise} developed a mixed noise removal model for HSIs by combining low-rank Tucker decomposition with $l_0$-norm-based regularizers, effectively capturing global correlations and improving sparse noise removal performance. In the context of TR decomposition, \citet{TR_superresolution} proposed a high-order coupled TR representation model for hyperspectral image super-resolution, which leverages shared latent core tensors and graph-Laplacian regularization to preserve spectral information. \citet{TR_fusion} designed a hyperspectral image fusion model based on TR decomposition, integrating logarithmic tensor nuclear norm regularization and weighted total variation to enhance spatial-spectral continuity. \citet{TR_resconstraction} proposed a subspace-based hyperspectral image reconstruction model that utilizes spectral quadratic variation regularized autoweighted TR decomposition.

{With the rapid advancement of deep learning, it has been widely applied in HSI processing. Convolutional neural networks (CNNs) have been utilized to extract spatial contextual information from HSIs using 2D or 3D convolutional kernels \citep{CNN_hong,CNN_hong2,CNN3,CNN5,DSCNet,2024_ReS3-ConvSet}. For example, \citet{CNN5} introduced a separable deep graph convolutional network that combines CNNs with prototype learning to enhance discriminative feature extraction. More recently, transformers have emerged as a powerful tool for hyperspectral image classification \citep{Transformer1,SpectralFormer,Transformer3,transfomer_hong,Transformer5}. For instance, \citet{SpectralFormer} proposed a transformer-based method that integrates a group-wise spectral embedding module and a cross-layer adaptive fusion module. Beyond conventional architectures, \citet{LRR-Net} combined low-rank representation with deep learning techniques, enabling automatic parameter learning for hyperspectral anomaly detection. \citet{SpectralGPT} proposed the first spectral remote sensing foundation model using a 3D generative pretrained transformer, incorporating million-scale progressive training and spectral-spatial coupling via tensor masking and multi-target reconstruction.}

\section{Proposed Method}\label{sec:proposed-method}
While tensor-based methods can effectively preserve the spatial structure of HSIs, they can only be applied to regular 3D data cubes, overlooking the fact that the same material in an HSI is usually distributed within irregular homogeneous regions. To solve this problem, we propose a novel tensor low-rank representation model that can process irregular data cubes and accordingly promote the representation ability.

Fig. \ref{mind} illustrates the framework of the proposed method. First, we use a typical superpixel segmentation method to divide the origin data into several irregular 3D cubes. Then we complete the irregular 3D data cubes with black shape-complementary data cubes to constitute regular tensor patches. Afterward, we design a novel low-rank representation model with a non-convex tensor norm to pursue the low-rank representation of the 3D patches that only takes the original irregular 3D data cube into account. To enhance the overall discriminative ability, we introduce a global negative nuclear norm term. The details of the proposed method are given below.

\begin{figure}
    \centering
    \includegraphics[width=0.9\linewidth]{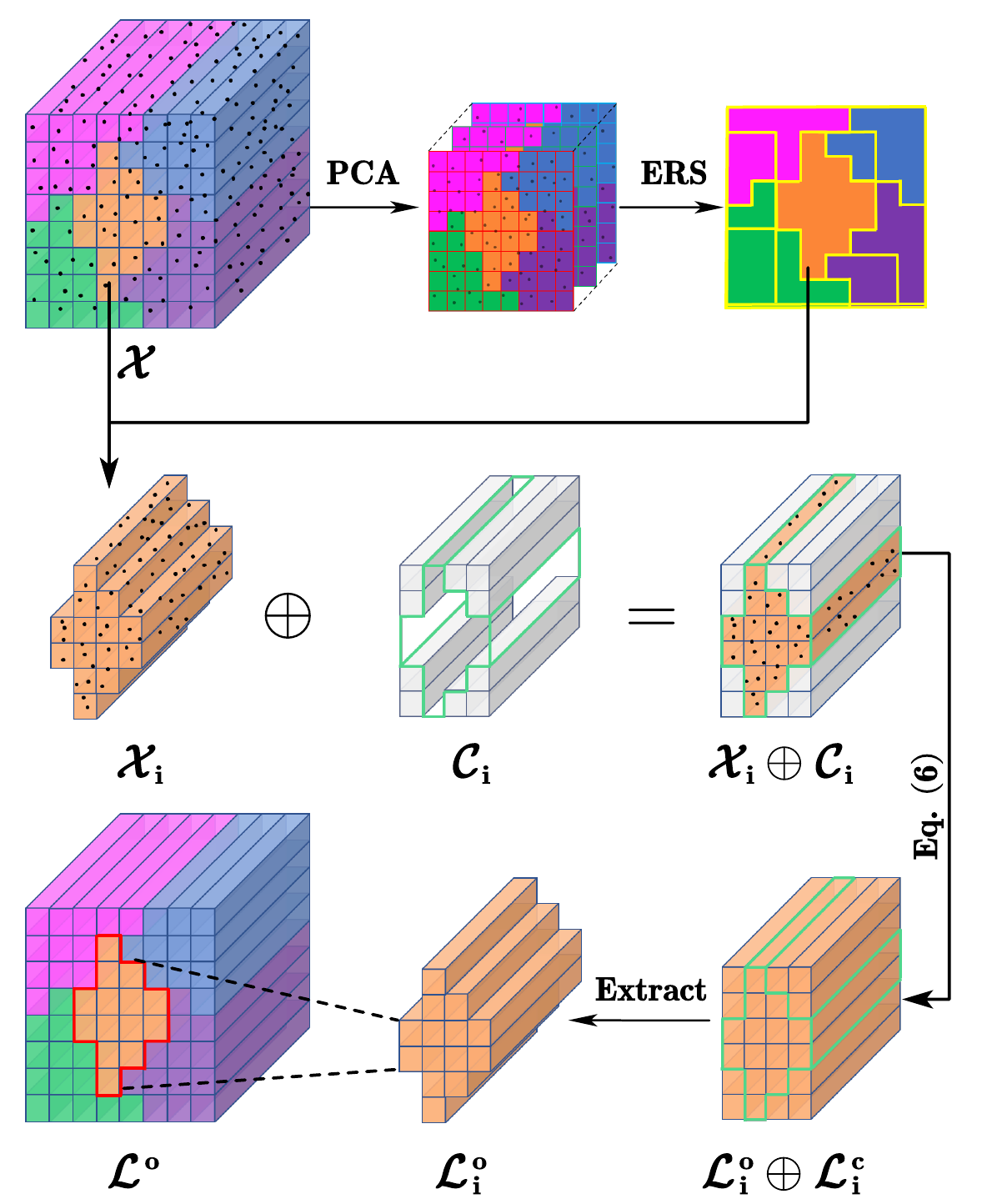}
    \caption{Flowchart for the low-rank approximation of irregular tensors.}
    \label{fig_core_idea}
\end{figure}

\subsection{Low-rank Representation for Irregular Tensor with A Non-convex Tensor Norm}

As depicted in Fig. \ref{fig_core_idea}, we define $\mathcal{X} \in \mathbb{R}^{n_1 \times n_2 \times n_3}$ as the input HSI, which usually suffers from spectral variations. The regions that have undergone spectral alterations are indicated by the black dots within the figure. Following \citep{superPCA}, we initially apply principal component analysis (PCA) to reduce the original dataset to dimensions of \( n_1 \times n_2 \times 3 \) and adopt entropy rate superpixel segmentation (ERS) \citep{ERS} to segment the compressed PCA data. By utilizing the segmentation results as indices, we can divide the original data into multiple non-overlapping irregular 3D cubes, i.e., $\mathcal{X}=\mathcal{X}_1 \oplus \mathcal{X}_2\oplus \cdots \mathcal{X}_i \oplus \cdots \oplus \mathcal{X}_n$, where $n$ is the number of superpixels and $\mathcal{X}_i$ is denoted for the \(i\)-th irregular tensor.

Although $\mathcal{X}_i |_{i=1}^n$ can capture irregular distributions of the input, we cannot directly apply the existing tensor low-rank norms on them to relieve the spectral variation problem. To this end, we introduce a complementary irregular tensor $\mathcal{C}_i $ to fill in the blanks. Specifically, we determine the smallest regular tensor with dimensions \( \mathbb{R}^{w_i \times b_i \times n_3} \) that can exactly accommodate $\mathcal{X}_i$. The complementary irregular tensor $\mathcal{C}_{i}$ is the region of the regular tensor that is not part of the original irregular tensor. Through introducing $\mathcal{C}_i$, we obtain a combined regular tensor $\mathcal{X}_i \oplus \mathcal{C}_i \in \mathbb{R}^{w_i \times b_i \times n_3} $, where we can perform low-rank approximation on it to extract low-rank representations corresponding to the original irregular tensor.

To make the low-rank approximation only affected by the original 3D cubes $\mathcal{X}_i$, not by the complementary 3D cube $\mathcal{C}_i$, we define the following notations. Let $\mathcal{L}_i^o$ and $\mathcal{S}_i^o$ represent the low-rank and sparse components of $\mathcal{X}_i$, respectively, i.e., $\mathcal{X}^i=\mathcal{L}_i^o+\mathcal{S}_i^o$. Similarly, we define $\mathcal{L}_i^c$ and $\mathcal{S}_i^c$ as the low-rank and sparse components of the complementary tensor $\mathcal{C}_i$, i.e., $\mathcal{C}^i=\mathcal{L}_i^c+\mathcal{S}_i^c$. Then, we apply tensor low-rank constraint on the combined regular tensor. i.e., on both $\mathcal{L}_i^o$ and $\mathcal{L}_i^c$ ($\mathcal{L}_i^o \oplus \mathcal{L}_i^c$), while different sparse constraints on $\mathcal{S}_i^o$ and $\mathcal{S}_i^c$, i.e., we only impose sparse constraint on $\mathcal{S}_i^o$, and impose no constraint on $\mathcal{S}_i^c$. This difference in constraints allows the sparse part of \(\mathcal{C}_i\) to remain free, while its low-rank part is guided by the low-rank component of \(\mathcal{X}_i\). As a result, the feature space of the low-rank part of \(\mathcal{C}_i\) will align with that of the low-rank part of \(\mathcal{X}_i\), thereby avoiding the influence of the complementary tensor \(\mathcal{C}_i\) on \(\mathcal{X}_i\), i.e., \textit{only the original irregular plays a role in low-rank tensor representation, while the added complementary irregular tensor does not affect the low-rank property of the original data}. Based on this approach, we formulate the following optimization model for the $i$-th irregular data cube:
\revice{\begin{equation}
\begin{aligned}
\min _{\{\mathcal{L}_i^o \oplus \mathcal{L}_i^c, \mathcal{S}_i^o\}} &\|\mathcal{L}_i^o \oplus \mathcal{L}_i^c\|_{*} + \lambda_i \|\mathcal{S}_i^o\|_{1}  \\
\text { s.t. } & \mathcal{X}^i=\mathcal{L}_i^o+\mathcal{S}_i^o, 
\end{aligned}
\label{Eq1}
\end{equation}}
where $\|\cdot\|_{*}$ is a tensor nuclear norm defined in Definition \ref{definition_nuclear_norm}, $\lambda_i $ is the regularization parameter of the $i$-th superpixel. As suggested by \citep{2020TRPCA}, we can set $\lambda_i = \alpha/\sqrt{\max(w_i, b_i)n_3}$ , where $\alpha$ is a variable related to tensor sparsity.

The tensor nuclear norm in Definition \ref{definition_nuclear_norm} regularizes all singular values of tensor data equally, which is not a perfect approximation of the rank function \citep{theorem4}. To achieve a better approximation of rank function, we introduce a non-convex tensor norm \citep{ETRPCA} for a tensor $\mathcal{A} \in \mathbb{R}^{n_1 \times n_2 \times n_3}$ as defined as:
\begin{equation}
\begin{aligned}
\|\mathcal{A}\|_{S_p}^p = \sum_{i=1}^{n_3} \|{\overline{\mathbf{A}}}^{(i)}\|_{S_p}^p = \sum_{i=1}^{n_3} \sum_{j=1}^{l} \sigma_{j}(\overline{{\mathbf{A}}}^{(i)})^{p}.
\end{aligned}
\label{non-convex norm}
\end{equation}
Here, $p (0 < p \leq 1)$ is the power of the singular value and $l = \min(n_1, n_2)$. $\overline{\mathbf{A}}^{(i)} \in \mathbb{R}^{n_1 \times n_2 }$ is the $i$-th frontal slice of $\overline{\mathcal{A}}\in \mathbb{R}^{n_1 \times n_2 \times n_3}$, where $\overline{\mathcal{A}} = fft(\mathcal{A}, [], 3)$. \revice{When $p$ is equal to 1, Eq. (\ref{non-convex norm}) is equivalent to the tensor nuclear norm defined in Definition \ref{definition_nuclear_norm}. For $p$ smaller than 1, the norm becomes nonconvex. In the limit as \( p \to 0 \), it approximates the tensor rank.} So the low-rank representation for irregular tensor with a non-convex tensor norm for the input tensor $\mathcal{X}$ is formulated as follows:{
\revice{\begin{equation}
\begin{aligned}
\min_{\left\{\mathcal{L}_i^o \oplus \mathcal{L}_i^c, \mathcal{S}_i^o\right\}_{i=1}^{n}} &\sum_{i=1}^n (\|\mathcal{L}_i^o \oplus \mathcal{L}_i^c\|_{S_p}^p + \lambda_i \|\mathcal{S}_i^o\|_{1} )\\
\qquad\text{s.t. } \; &\mathcal{X} = \mathcal{L}^o + \mathcal{S}^o,
\end{aligned}
\label{EQ_combined}
\end{equation}}
where $\mathcal{X} = \mathcal{X}_1 \oplus \mathcal{X}_2 \oplus \dotsc \oplus \mathcal{X}_n$,  
$\mathcal{L}^o = \mathcal{L}^o_1 \oplus \mathcal{L}^o_2 \oplus \dotsc \oplus \mathcal{L}^o_n$ and $\mathcal{S}^o = \mathcal{S}^o_1 \oplus \mathcal{S}^o_2 \oplus \dotsc \oplus \mathcal{S}^o_n$.}

\subsection{Promoting the Discriminative Ability}

However, the above model processes the irregular 3D data cubes individually, overlooking that a superpixel potentially contains more than one category of materials as shown in Fig. \ref{mind} (II). When seeking low-rank representation in the local area, the non-dominant class in one superpixel block will be over-penalized to make it close to the dominant one, which leads to a decrease in discriminability between different categories in one superpixel and a decrease in similarity among pixels belonging to the same category in different superpixels. For example, in Fig. \ref{mind} (II), we can see that blocks $B$, $E_1$ and $E_2$ are the non-dominant classes within each superpixel and blocks $E_1$ and $E_2$ originate from one category in the original data. But after executing Eq. (\ref{EQ_combined}), they are over-penalized, thereby pushed closer to the categories that occupy the dominant position in irregular 3D data cubes as shown in Fig. \ref{mind} (IV). This will decrease the discriminability among different categories in one superpixel (e.g., $A$ and $B$, $E_1$ and $C$, $E_2$ and $D$). And the similarity of the pixels belonging to one category also decreases (e.g., $E_1$ and $E_2$). 

To address this issue, we introduce a global regularization term $-\|\mathcal{L}^o\|_{\star}:=-\|\mathbf{L}_{(3)}\|_*$, where $\mathbf{L}_{(3)} \in \mathbb{R}^{n_1  n_2 \times n_3}$ is the third-order unfolding $\mathcal{L}^o$ and $\|\cdot\|_*$ is the nuclear norm of the 2D matrix. 
Different from the irregular tensor low-rank prior, the global term aims to promote discriminability among different categories and improve the consistency of the same category in different superpixels by increasing the singular values of the whole matrix. The final objective function of ITLRR is formulated as:\revice{
\begin{align}
\min_{\left\{\mathcal{L}_i^o \oplus \mathcal{L}_i^c, \mathcal{S}_i^o\right\}_{i=1}^{n}} 
& \sum_{i=1}^n \left( \|\mathcal{L}_i^o \oplus \mathcal{L}_i^c\|_{S_p}^p + \lambda_i \|\mathcal{S}_i^o\|_{1} \right) - \beta \|\mathcal{L}^o\|_{\star} \nonumber \\
&\text{s.t. }  \mathcal{X} = \mathcal{L}^o + \mathcal{S}^o, \label{EQ3}
\end{align}}
where $\beta \geq 0$ serves as a trade-off parameter, $\mathcal{X} = \mathcal{X}_1 \oplus \mathcal{X}_2 \oplus \dotsc \oplus \mathcal{X}_n$,  
$\mathcal{L}^o = \mathcal{L}^o_1 \oplus \mathcal{L}^o_2 \oplus \dotsc \oplus \mathcal{L}^o_n$ and $\mathcal{S}^o = \mathcal{S}^o_1 \oplus \mathcal{S}^o_2 \oplus \dotsc \oplus \mathcal{S}^o_n$. By balancing the global term and the local irregular low-rank term, the dissimilarity among pixels belonging to different categories is expected to increase, while that of the pixels belonging to the same category will decrease.

\subsection{Optimization}

To solve the optimization problem in Eq. (\ref{EQ3}), we first derive its augmented Lagrangian form:
\begin{equation}
\resizebox{0.8\hsize}{!}{$
\begin{aligned}
\min _{\left\{\mathcal{L}_i^o \oplus \mathcal{L}_i^c, \mathcal{S}_i^o\right\}_{i=1}^{n}}&\sum_{i=1}^n \left(\|\mathcal{L}^o_i \oplus \mathcal{L}^c_i\|_{S_p}^p+\lambda_i\|\mathcal{S}^o_i\|_{1}\right)
-\beta\|\mathcal{L}^o\|_{\star}\\
&+\frac{\mu}{2}\left\|\mathcal{L}^o-\left(\mathcal{X}-\mathcal{S}^o+\frac{\mathcal{Y}}{\mu}\right)\right\|_{F}^{2},
\end{aligned}$}
\label{lagrangin}
\end{equation}
where $\mathcal{Y} = \mathcal{Y}_1 \oplus \mathcal{Y}_2 \oplus \dotsc \oplus \mathcal{Y}_n$, $\mathcal{Y}_i \in \mathbb{R}^{w_i \times b_i \times n_3}$ serves as the Lagrangian multiplier for the $i$-th superpixel, and $\mu > 0$ acts as the penalty parameter. Note that there are no constraints on $\mathcal{C}_i$ and $\mathcal{S}^c_i$, and we only need to solve for  $\mathcal{L}_i^o \oplus \mathcal{L}_i^c$ and $\mathcal{S}^o$. We solve Eq. (\ref{lagrangin}) in an alternative manner, i.e., alternatively update one variable and fix others.

1) The $\left\{\mathcal{L}_i^o \oplus \mathcal{L}_i^c\right\}_{i=1}^{n}$ sub-problem can be expressed as:
\begin{equation}
\begin{aligned}
\underset{\left\{\mathcal{L}_i^o \oplus \mathcal{L}_i^c\right\}_{i=1}^{n}}{\operatorname{min}}  \hspace{0.5em} &\sum_{i=1}^n \|\mathcal{L}^o_i \oplus \mathcal{L}^c_i\|_{S_p}^p-\beta\|\mathcal{L}^o\|_{\star}\\
&+\frac{\mu}{2}\left\|\mathcal{L}^o-\left(\mathcal{X}-\mathcal{S}^o+\frac{\mathcal{Y}}{\mu}\right)\right\|_{F}^{2},
\end{aligned}
\label{ysub}
\end{equation}
where the second term is concave, making it challenging to solve.  In order to simplify Eq. (\ref{ysub}), we first linearize the concave term \(-\|\mathcal{L}^o\|_{\star}\) through its first-order Taylor expansion around \(\mathcal{L}^{o^{t}}\), which was obtained in the preceding iteration, with \(t \geq 0\) denoting the iteration index. The approximation is expressed as follows:
\begin{equation}
-\|\mathcal{L}^o\|_{\star} \approx -\left\|\mathcal{L}^{o^{t}}\right\|_{\star} - \left\langle \mathcal{T}^{t}, \mathcal{L}^o- \mathcal{L}^{o^{t}} \right\rangle,
\label{linear}
\end{equation}
where \(\mathcal{T}^{t} \in \mathbb{R}^{n_1 \times n_2 \times n_3}\) represents the sub-gradient of \(\|\mathcal{L}^o\|_{\star}\) \citep{-lowrank} evaluated at \(\mathcal{L}^{o^{t}}\), defined as \(\mathcal{T}^{t} = \partial \|\mathcal{L}^{o^{t}}\|_{\star} := \mathscr{F}_3(\partial\|\mathbf{L}_{(3)}^{o^{t}}\|_{*})\). The explicit formulation of \(\partial\|\cdot\|_{*}\) is detailed in Theorem \ref{linearize}.

With Eq. (\ref{linear}), we can approximate Eq. (\ref{ysub}) as follows:
\begin{equation}
\resizebox{0.8\hsize}{!}{$
\begin{aligned}
&\min _{\left\{\mathcal{L}_i^o \oplus \mathcal{L}_i^c\right\}_{i=1}^{n}} g(\left\{\mathcal{L}_i^o \oplus \mathcal{L}_i^c\right\}_{i=1}^{n}) \\
&\qquad=\sum_{i=1}^n \|\mathcal{L}^o_i \oplus \mathcal{L}^c_i\|_{S_p}^p-\beta\|\mathcal{L}^{o^{t}}\|_{\star}-\beta\langle\mathcal{T}^{o^{t}},\mathcal{L}^o-\mathcal{L}^{o^{t}}\rangle \\
&\qquad\quad+\frac{\mu}{2}\left\|\mathcal{L}^o-\left(\mathcal{X}-\mathcal{S}^o+\frac{\mathcal{Y}}{\mu}\right)\right\|_{F}^{2}\\
&\qquad=\sum_{i=1}^n \|\mathcal{L}^o_i \oplus \mathcal{L}^c_i\|_{S_p}^p+\frac{\mu}{2}\|\mathcal{L}^o-\mathcal{A}\|_{F}^{2}\\
&\qquad\quad-\beta\|\mathcal{L}^{o^{t}}\|_{\star}
-\beta\langle\mathcal{T}^{o^{t}},-\mathcal{L}^{o^{t}}\rangle,
\end{aligned}
$}
\label{yend}
\end{equation}
where  $\mathcal{A}=\mathcal{X}-\mathcal{S}^o+\frac{\mathcal{Y}+\beta\mathcal{T}}{\mu}$ and the term $-\beta\|\mathcal{L}^{o^{t}}\|_{\star}
-\beta\langle\mathcal{T}^{o^{t}},-\mathcal{L}^{o^{t}}\rangle$ is a constant irrelevant to $\mathcal{L}^o$.

Since $\left\{\mathcal{L}_i^o \oplus \mathcal{L}_i^c\right\}_{i=1}^{n}$ are independent, Eq. (\ref{yend}) can be divided into $n$ independent sub-problems. The sub-problem for the $i$-th superpixel formulated as:
\begin{equation}
\begin{aligned}
\min _{\mathcal{L}_i^o \oplus \mathcal{L}_i^c}
g(\mathcal{L}_i^o \oplus \mathcal{L}_i^c)
&=\|\mathcal{L}^o_i \oplus \mathcal{L}^c_i\|_{S_p}^p
+\frac{\mu}{2}\|\mathcal{L}_i^o-\mathcal{A}_{i}\|_{F}^{2}\\
&=\|\mathcal{L}^o_i \oplus \mathcal{L}^c_i\|_{S_p}^p\\
&\quad+\frac{\mu}{2}\|\mathcal{L}^o_i \oplus \mathcal{L}^c_i-(\mathcal{A}_{i} \oplus \mathcal{L}^c_i)\|_{F}^{2}.
\label{l}
\end{aligned}
\end{equation}
Eq. (\ref{l}) can be simplified to:
\begin{equation}
\min _{\mathcal{N}_i}\|\mathcal{N}_i\|_{S_p}^p
+\frac{\mu}{2}\|\mathcal{N}_i-\mathcal{M}_i\|_{F}^{2},
\label{22}
\end{equation}
where $\mathcal{N}_i= \mathcal{L}_i^o \oplus \mathcal{L}_i^c$ and $\mathcal{M}_i = \mathcal{L}^c_i \oplus 
\mathcal{A}_{i}$.

In the Fourier domain, Eq. (\ref{22}) can be expressed as:
\begin{equation}
\begin{aligned}
\min _{\mathcal{N}_i}
g(\mathcal{N}_i)
&=\sum_{j=1}^{n_3}\|\overline{\mathbf{N}}_i^{(j)}\|_{S_p}^p
+\frac{\mu}{2}\|\overline{\mathcal{N}_i}-\overline{\mathcal{M}_i}\|_{F}^{2}\\
&=\sum_{j=1}^{n_3}\left(\|\overline{\mathbf{N}}_i^{(j)}\|_{S_p}^p
+\frac{\mu}{2}\|\overline{\mathbf{N}}_i^{(j)}
-\overline{\mathbf{M}}_i^{(j)}\|_{F}^{2}\right).
\label{l2}
\end{aligned}
\end{equation}
In Eq. (\ref{l2}), each variable $\overline{\mathbf{N}}_i^j$ is independent. Therefore, Eq. (\ref{l2}) can be written as:
\begin{equation}
\min _{\overline{\mathbf{N}}_i^{(j)}}
\|\overline{\mathbf{N}}_i^{(j)}\|_{S_p}^p
+\frac{\mu}{2}\|\overline{\mathbf{N}}_i^{(j)}
-\overline{\mathbf{M}}_i^{(j)}\|_{F}^{2} .
\label{l sub-problem}
\end{equation}

According to Theorem \ref{optial}, the optimal solution to Eq. (\ref{l sub-problem}) is given by:
\begin{equation} 
\overline{\mathbf{N}}_i^{(j)*}=\mathbf{U} \mathbf{S}_{\epsilon} \mathbf{V}^{T},
\label{l_solution}
\end{equation}
where $\overline{\mathbf{M}}_i^{(j)}=\mathbf{U} \mathbf{S} \mathbf{V}^{T}$ and \small $\mathbf{S}_{\epsilon}=$ $\operatorname{Diag}\{(\mathbf{S}_{i i}- {p(\sigma_{i}(\overline{\mathbf{N}}_i^j))^{p-1}}/{\mu})_{+}\}$ \normalsize.

2) The $\left\{\mathcal{S}_i^o\right\}_{i=1}^{n}$ sub-problem can be expressed as follows:
\begin{equation}
\sum_{i=1}^{n}\left(\underset{\mathcal{S}_i^o}{\operatorname{min}} \lambda_i\|\mathcal{S}^o_i\|_{1} +\frac{\mu}{2}\left\|\mathcal{S}^o_i-\mathcal{B}_i\right\|_{F}^{2}\right),
\label{S_problem_last}
\end{equation}
where $\mathcal{B}_i=\mathcal{X}_i-\mathcal{L}^o_i+{\mathcal{Y}_i}/{\mu}$. 

Eq. (\ref{S_problem_last}) is a set of $\ell_1$ norm minimization problems, which can be solved by the soft-thresholding operator, i.e., 
\begin{equation}    
\mathcal{S}^{o^*}_i= \mathrm{T}_{{\lambda_i}/{\mu}}\left(\mathcal{B}_i\right),
\label{S_lotution}
\end{equation}
where the $(i, j, k)_{th}$ element of $\mathrm{T}_{{\lambda_i}/{\mu}}\left(\mathcal{B}_i\right)$ is defined as follows:
\small
\begin{equation}
(\mathrm{T}_{{\lambda_i}/{\mu}}\left(\mathcal{B}_i\right))_{ijk} = \operatorname{sign}((\mathcal{B}_i)_{ijk}) \cdot \max (|(\mathcal{B}_i)_{ijk}| - {\lambda_i}/{\mu}, 0).
\end{equation}
\normalsize

3) The Lagrangian parameter and the penalty  are updated in each iteration as:
\begin{equation}
\left\{\begin{array}{l}
\mathcal{Y}^{t+1}_i=\mathcal{Y}^{t}_i+\mu_{i}(\mathcal{X}_i-\mathcal{L}_{i}^{o^{t+1}}-\mathcal{S}_{i}^{o^{t+1}}) \\
\mu^{t+1}=\min \left(\rho \mu^{t}, \mu_{\max }\right)
\end{array}\right..
\label{parameters}
\end{equation}
The overall optimization procedure is summarized in Algorithm \ref{alg:algorithm1}. 

\begin{algorithm}
	\caption{{Solve Eq. (\ref{EQ3}) by ADMM \citep{cite_admm}}.}
	\label{alg:algorithm1}
        \KwIn{\text{tensor data} $\mathcal{X}$, \text{parameters} $\lambda$ and $\beta$, number of superpixels $n$.}

        Initialize $\rho=1.1$, $\mu=1e-10$, $\mu_{\text{max}}=1e10$ and 

        $\epsilon=1e-3$.
	 
        Split the input data $\mathcal{X}$ into $n$ irregular tensors $\mathcal{X}_i$.

        Create $\mathcal{L}_i^o$, $\mathcal{L}_i^c$, $\mathcal{S}_i^o$ and $\mathcal{Y}_i$ with dimensions equivalent 
        
        to $\mathcal{X}_i$.

	\While{\textnormal{not converged}}{
		Perform linearization as shown in Eq. (\ref{linear});
		
		Update $\mathcal{L}_i^o \oplus \mathcal{L}_i^c$ using Eq. (\ref{l_solution});
		
		Update $\mathcal{S}^o$ using Eq. (\ref{S_lotution});
		
		Update $\mathcal{Y}$ and $\mu$ according to Eq. (\ref{parameters});
	 
		Check the convergence metric: $\mathrm{error} = \max(\|\mathcal{L}^{o^{t+1}} - \mathcal{L}^{o^{t}}\|_\infty, \|\mathcal{S}^{o^{t+1}} - \mathcal{S}^{o^{t}}\|_\infty, \|\mathcal{X} - \mathcal{L}^{o^{t+1}} - \mathcal{S}^{o^{t+1}}\|_\infty) $
		
		\If{$\mathrm{error} \le \epsilon$}{
			break;
		}
	}

	\KwOut{\text{low-rank tensor} $\mathcal{L}^o$.}  
\end{algorithm}

\section{Experiments}\label{sec:experiments}

\subsection{Datasets}

In this section, we use four datasets\footnote{The first three datasets (\textit{Indian Pines}, \textit{Salinas} and \textit{Pavia University}) are obtained from the following website: \url{https://www.ehu.eus/ccwintco/index.php?title=Hyperspectral_Remote_Sensing_Scenes}. The last dataset (\textit{WHU-Hi-LongKou}) is obtained from \url{http://rsidea.whu.edu.cn/resource_WHUHi_sharing.htm}.} to evaluate the effectiveness of the proposed method. The details of the datasets are as follows:

\begin{enumerate}
\item{\textit{Indian Pines}: This scene was acquired by NASA’s Airborne Visible/Infrared Imaging Spectrometer (AVIRIS) sensor over West Lafayette, IN, USA, on June 12, 1992, which contains 145 × 145 pixels with 200 bands.}

\item{\textit{Salinas}: This scene was acquired by the AVIRIS sensor over Salinas Valley, CA, USA, which contains 512 × 217 pixels with 204 bands.}

\item{\textit{Pavia University}: This scene was acquired by the Reflective Optics System Imaging Spectrometer (ROSIS) sensor over  Italy, which consists of 610 × 340 pixels with 103 bands.}

\item{\textit{WHU-Hi-LongKou} \citep{WH_dataset}\citep{WH_dataset2}: This scene was acquired over Longkou Town, Hubei province, China, on July 17, 2018, which consists of 550 × 400 pixels with 270 bands. A rectangular part (from 151 to 320 rows and 51 to 300 columns) suffering from noise heavily is used for testing.}
\end{enumerate}

\begin{table}[ht]
    \centering
    \caption{Parameter Settings of Our Method on Different Datasets}
    \begin{tabular}{ccccc}
        \toprule    
        &Indian Pines&Salinas&Pavia University   &{WHU-Hi-LongKou}\\
        \hline
        $p$&0.1 &0.1 &0.1 &0.7\\
        $n$&30 &20 &10 &10\\
        $\alpha$&1e-7&1e-6&5e-6&5e-4\\
        $\beta$&1e-5&1e-2&1e-6 &1e-5\\
        \bottomrule
    \end{tabular}
    \label{parameters_opt}
\end{table}

\begin{figure*}
\centering
 \begin{minipage}{4cm}
 \centering
  \includegraphics[width=1.1\textwidth]{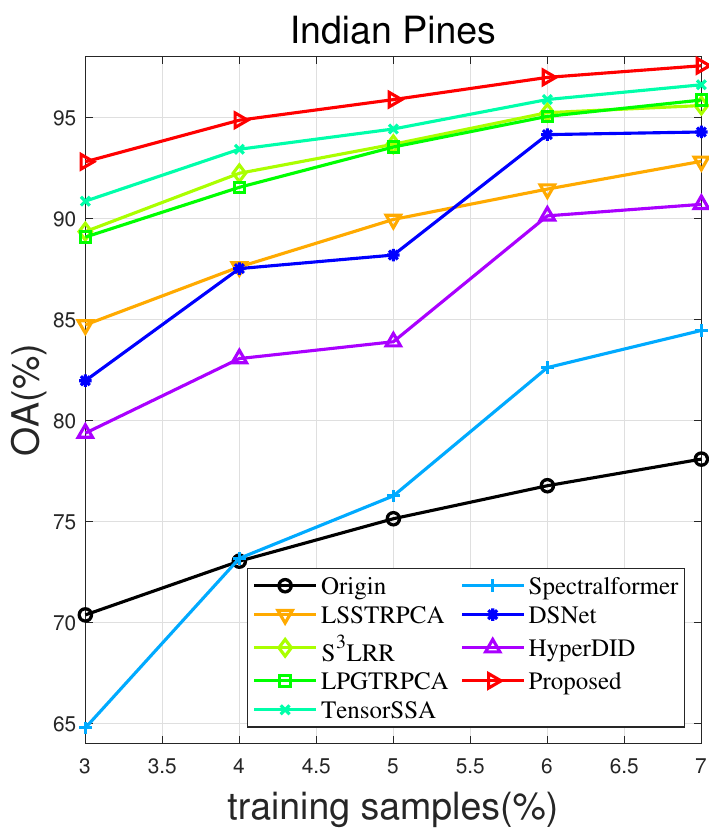}
 
  \end{minipage}\hspace{5mm}
  \begin{minipage}{4cm}
  \centering
  \includegraphics[width=1.1\textwidth]{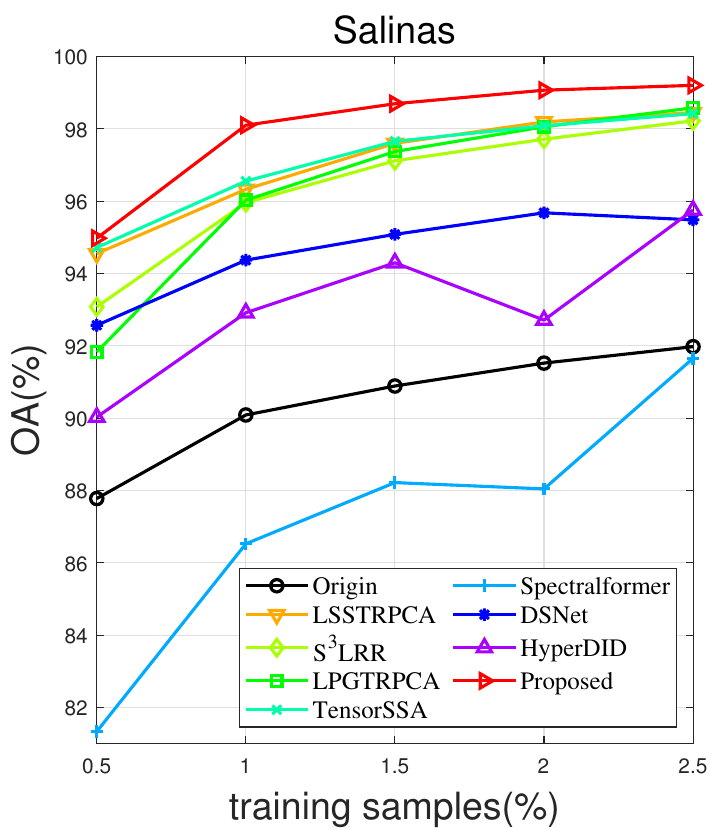}

    \end{minipage}\hspace{5mm}
  \begin{minipage}{4cm}
  \centering
  \includegraphics[width=1.1\textwidth]{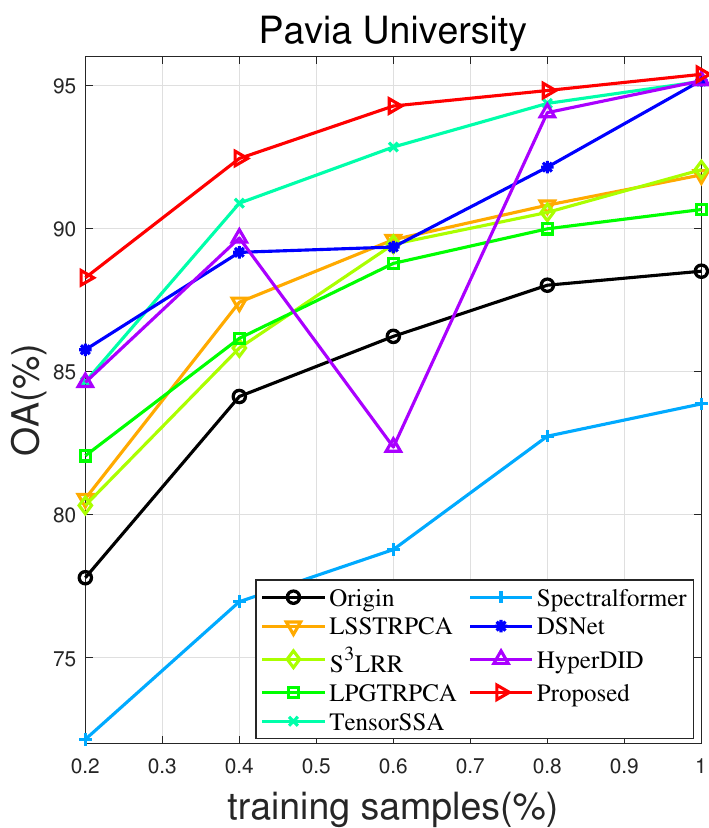}

    \end{minipage}\hspace{5mm}
  \begin{minipage}{4cm}
  \centering
  \includegraphics[width=1.1\textwidth]{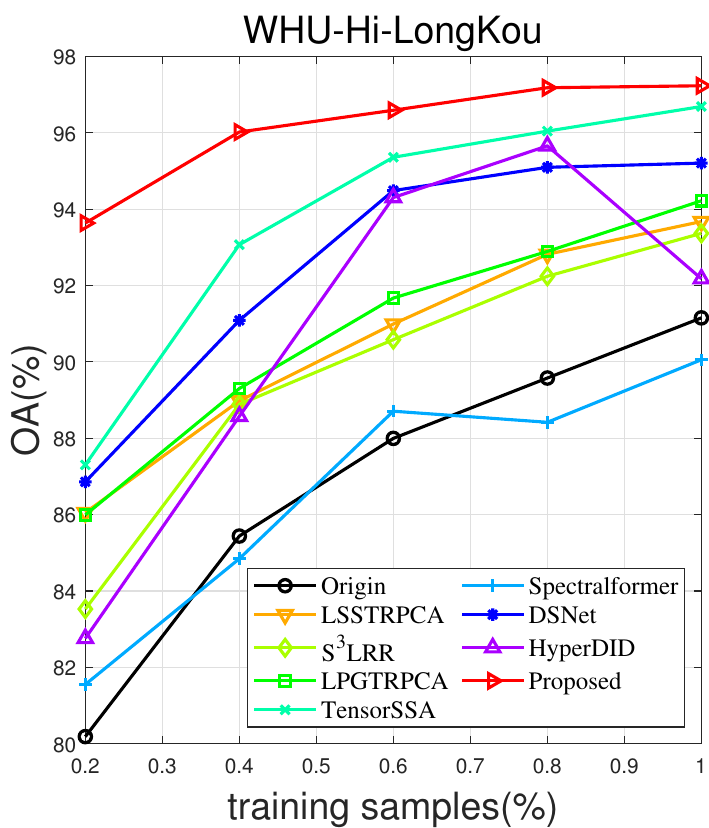}

  \end{minipage}
\caption{Comparison of the classification accuracy of different methods under various percentages of training samples on four datasets.}
  \label{compare-all}
\end{figure*}

\begin{figure*}
\centering
\subfigure[]{\includegraphics[height=1.69cm]{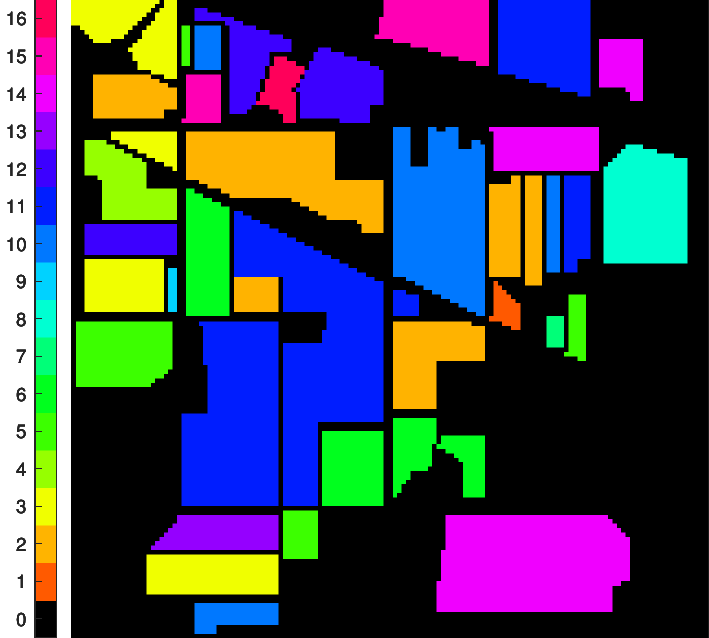}%
\label{subfig_Indian_gt_2}}
\subfigure[]{\includegraphics[height=1.69cm]{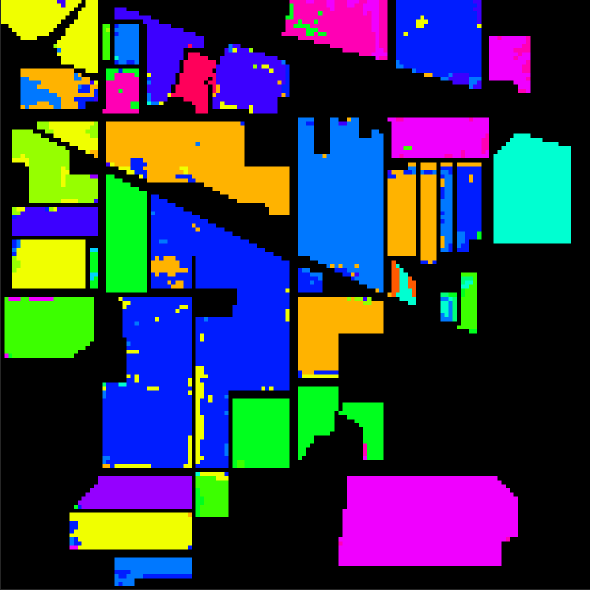}%
\label{subfig_Indian_spe}}
\subfigure[]{\includegraphics[height=1.69cm]{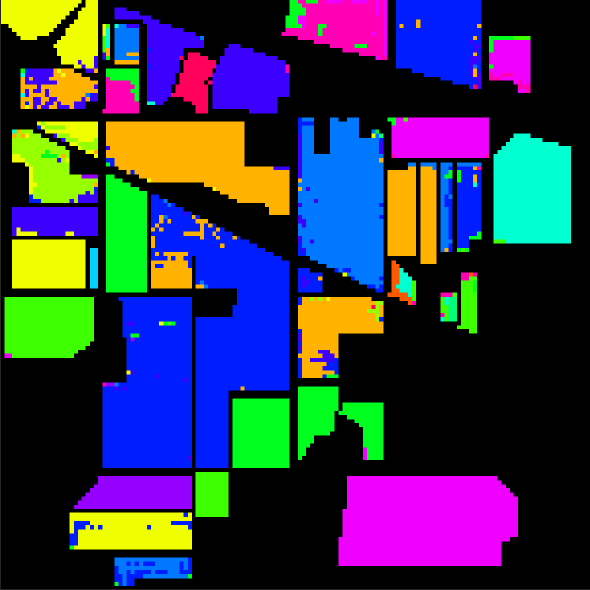}%
\label{subfig_Indian_DSNet}}
\subfigure[]{\includegraphics[height=1.69cm]{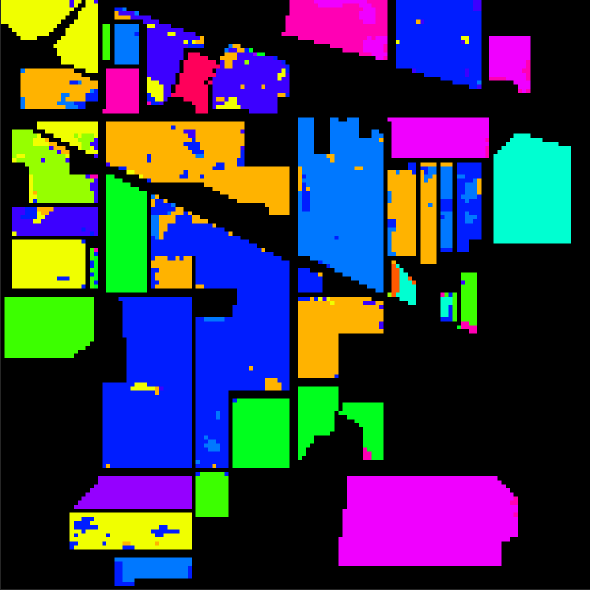}%
\label{subfig_Indian_HyperDID}}
\subfigure[]{\includegraphics[height=1.69cm]{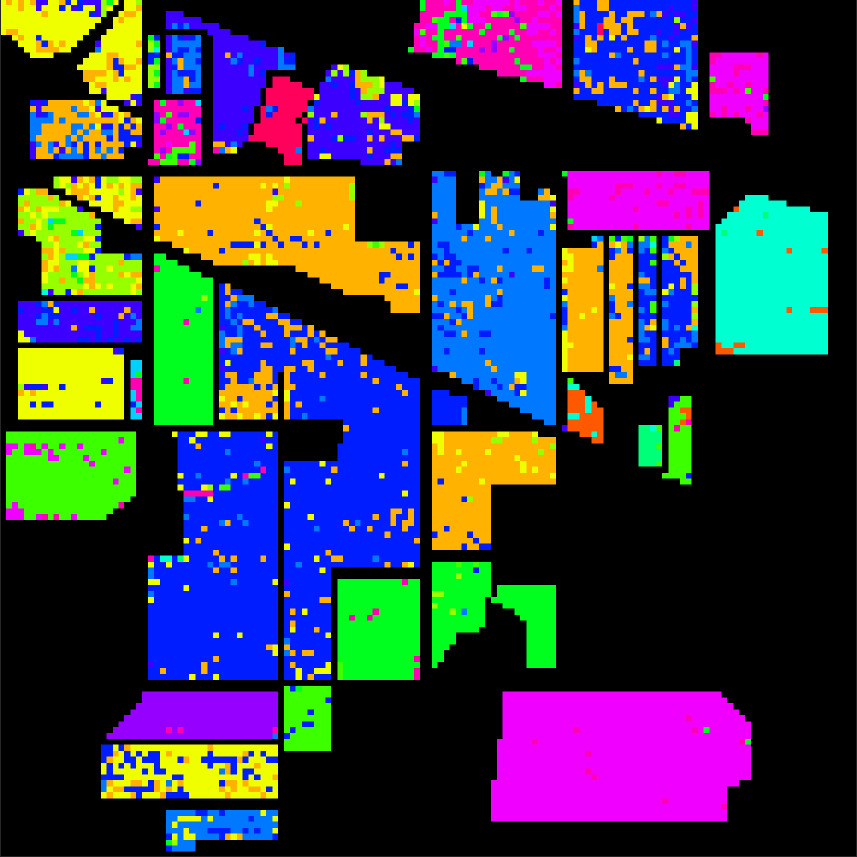}%
\label{subfig_Indian_Origin_2}}
\subfigure[]{\includegraphics[height=1.69cm]{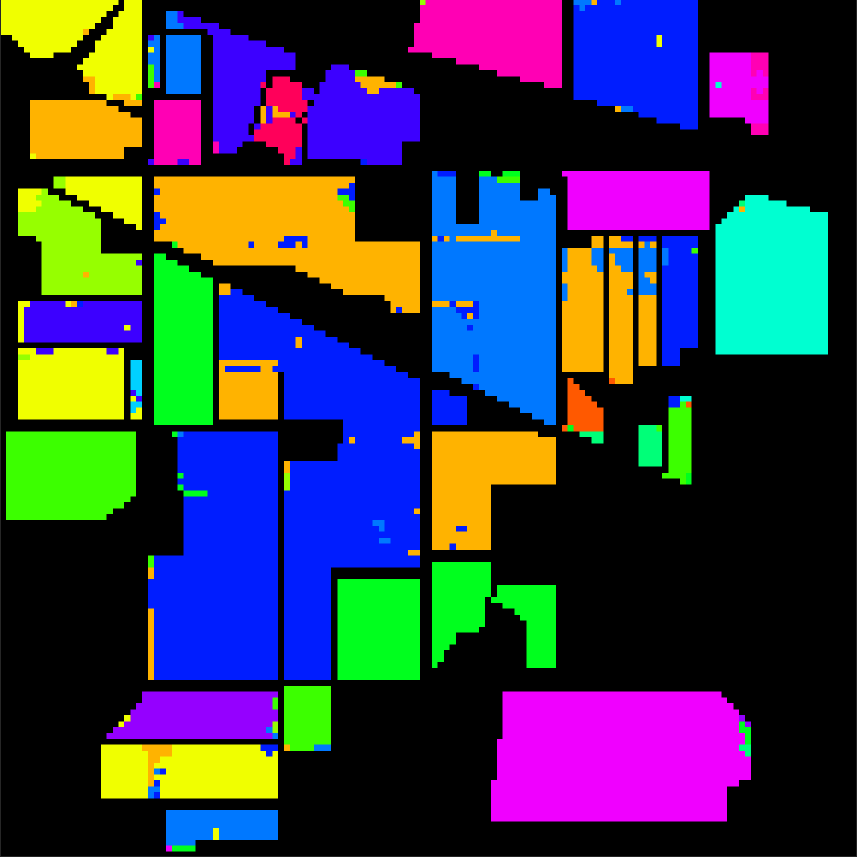}%
\label{subfig_Indian_LSSTRPCA}}
\subfigure[]{\includegraphics[height=1.69cm]{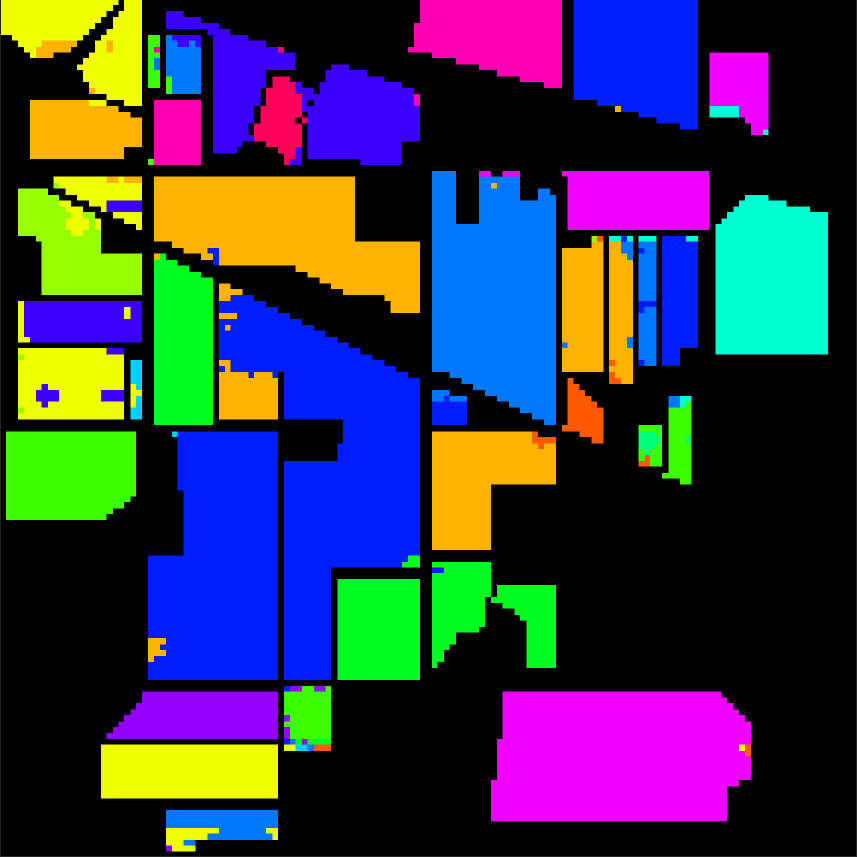}%
\label{subfig_Indian_S3LRR}}
\subfigure[]{\includegraphics[height=1.69cm]{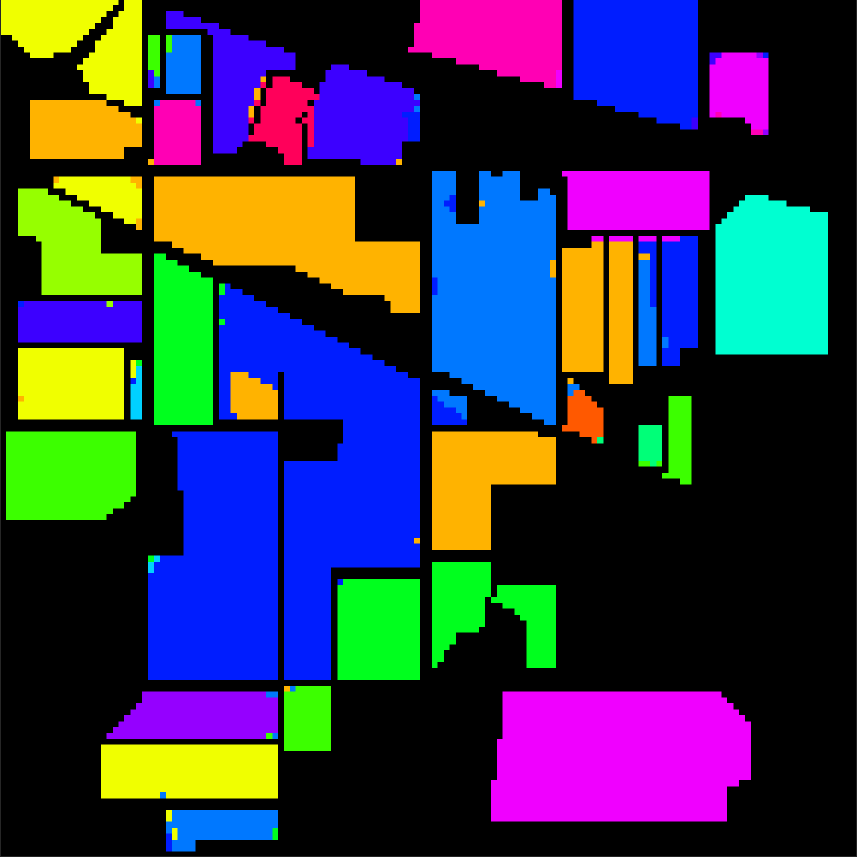}%
\label{subfig_Indian_LPGTRPCA}}
\subfigure[]{\includegraphics[height=1.69cm]{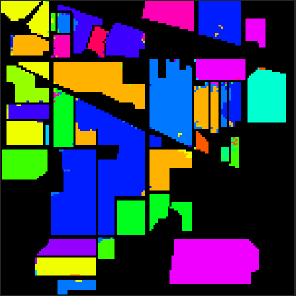}%
\label{subfig_Indian_SSA}}
\subfigure[]{\includegraphics[height=1.69cm]{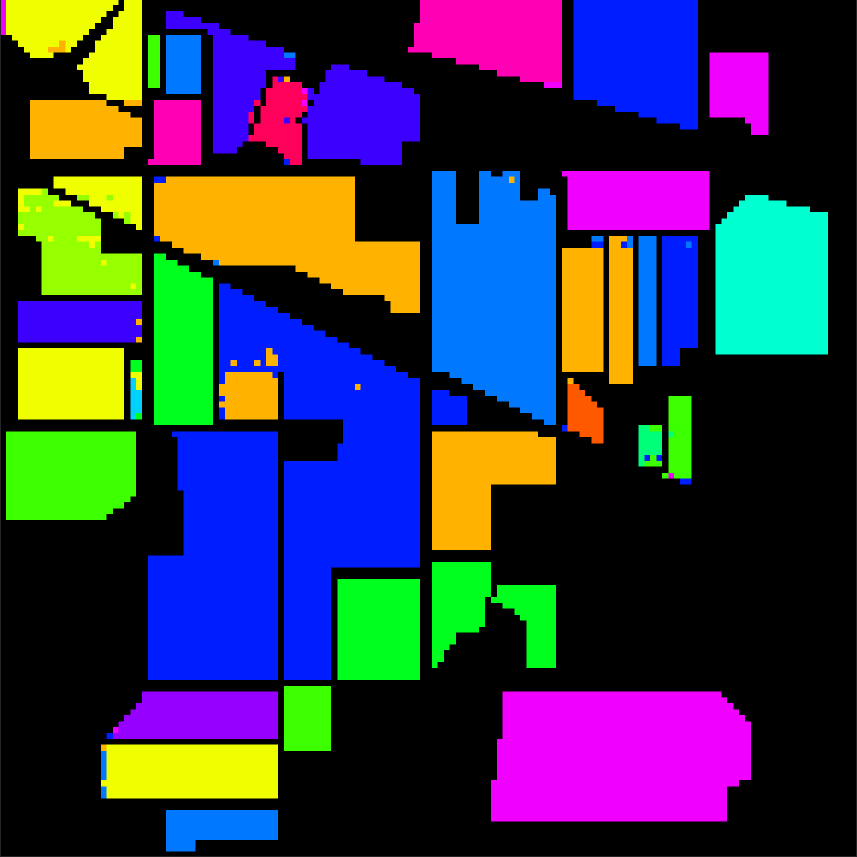}%
\label{subfig_Indian_proposed_2}}
\caption{Classification maps of comparison methods on \textit{Indian Pines} with 10\% training samples. (a) Groundtruth. (b)
Origin. (c) SpeFormer. (d) DSNet. (e) HyperDID. (f) LSSTRPCA. (g) $\mathrm{S^3LRR}$. (h) LPGTRPCA. (i) TensorSSA. (j) Proposed.}
\label{comparefig1}
\end{figure*}

\begin{table*}[ht]
\centering
\caption{Classification Performance of Comparison Methods on \textit{Indian Pines} with 10\% Training Samples. Optimal Values Are Denoted in Bold, and the Second-Best Values Are Underlined. $\bullet$/$\circ$ Indicates Whether the Performance of DSNet on Data Processed by Our Method Is Superior/Inferior to That on the Original Data.}
\scalebox{0.75}{
\begin{tabular}{c|cc|c|ccc|cccc|c:c} 
\hline\hline
\multicolumn{4}{c|}{} &\multicolumn{3}{c|}{Deep learning methods}&\multicolumn{4}{c|}{Tensor decomposition methods} &\multicolumn{2}{c}{Ours}\\  \hline 
\#  & Train&Test & Origin    & SpeFormer \citep{SpectralFormer} &  DSNet \citep{DSCNet} &HyperDID \citep{HyperDID}   & LSSTRPCA \citep{2020LSSTRPCA}    & $\mathrm{S^3LRR}$ \citep{2018S3LRR}        & LPGTRPCA \citep{2022LPGTRPCA} &TensorSSA \citep{2023TensorSSA} & Proposed               &DSNet+Ours\\\hline
1 & 5&41
& 63.75 ± 14.75
 & 40.00 ± 14.00
& 62.44 ± 20.15
& 8.78 ± 17.56
& 85.50 ± 9.23& 85.63 ± 12.35& 91.25 ± 9.41  &\underline{92.75 ± 5.84}
& \textbf{94.00 ± 1.37}
 &$\bullet$90.73 ± 10.39
\\
2 & 143&1285
& 79.19 ± 2.49
 & 88.95 ± 3.69
& 82.51 ± 10.93
& 94.23 ± 2.56
& 93.93 ± 2.09& \underline{97.25 ± 1.20} & 97.11 ± 1.22 &96.69 ± 1.12
& \textbf{98.72 ± 0.77}
 &$\bullet$98.54 ± 0.81
\\
3 & 83&747
& 71.03 ± 3.02
 & 88.15 ± 4.19
& 88.78 ± 4.72
& 93.04 ± 2.84
& 92.73 ± 3.16& 95.32 ± 2.19& 96.39 ± 1.61  &\textbf{96.92 ± 1.64}
& \underline{96.86 ± 1.38}
 &$\bullet$99.20 ± 1.10
\\
4 & 24&213
& 55.00 ± 6.32
 & 77.80 ± 6.30
& 68.26 ± 12.19
& 83.76 ± 8.62
& 91.22 ± 5.66& 94.23 ± 3.74& \textbf{97.58 ± 2.10} &\underline{96.92 ± 3.01}
& {94.46 ± 2.88}
 &$\bullet$97.84 ± 1.56
\\
5 & 49&434
& 90.18 ± 2.33
 & 92.97 ± 2.98
& 88.80 ± 4.60
& \underline{96.91 ± 1.43}
& 92.51 ± 2.61& 94.99 ± 2.09& {96.10 ± 2.44}  &95.02 ± 2.48
& \textbf{98.38 ± 1.02}
 &$\bullet$99.08 ± 0.73
\\
6 & 73&657
& 95.60 ± 1.50
 & 97.46 ± 2.26
& 99.33 ± 0.72
& 99.36 ± 0.51
& 99.32 ± 0.65& \underline{99.50 ± 0.46} & 99.02 ± 0.77 &99.18 ± 0.67
& \textbf{99.54 ± 1.02}
 &$\bullet$99.88 ± 0.18
\\
7 & 3&25
& 75.80 ± 12.34
 & 39.23 ± 18.90
& 63.20 ± 21.97
& 7.20 ± 14.40
& 71.00 ± 18.44& 70.80 ± 18.72& \textbf{96.20 ± 6.15} &\underline{91.80 ± 7.73}
& {88.00 ± 12.00}
 &$\bullet$64.80 ± 17.42
\\
8 & 48&430
& 96.84 ± 1.81
 & 99.12 ± 0.78
& 99.07 ± 1.03
& \underline{99.81 ± 0.37}
& 98.82 ± 1.79& 99.69 ± 0.75& {99.74 ± 0.49}  &99.24 ± 0.96
& \textbf{100.00 ± 0.00}
 &$\bullet$99.86 ± 0.28
\\
9 & 2&18
& 37.22 ± 13.50
 & 40.00 ± 25.56
& 43.33 ± 23.41
& 0.00 ± 0.00
& 56.67 ± 17.44& {71.67 ± 15.39} & 68.06 ± 22.43 &\textbf{96.94 ± 7.09}
& \underline{82.22 ± 18.17}
 &$\bullet$66.67 ± 22.22
\\
10 & 98&874
& 74.30 ± 2.69
 & 90.11 ± 4.29
& 93.00 ± 3.28
& 92.01 ± 3.06
& 91.21 ± 2.21& 94.43 ± 1.81& {95.93 ± 1.57}  &\underline{96.39 ± 1.75}
& \textbf{99.11 ± 0.83}
 &$\bullet$99.57 ± 0.29
\\
11 & 246&2209
& 79.41 ± 2.44
 & 94.32 ± 2.14
& 96.04 ± 2.21
& 95.27 ± 2.06
& 95.47 ± 1.15& {97.95 ± 0.82} & 97.69 ± 0.64 &\underline{98.75 ± 0.48}
& \textbf{99.38 ± 0.40}
 &$\bullet$99.33 ± 0.47
\\
12 & 60&533
& 70.37 ± 3.26
 & 81.65 ± 4.74
& 86.04 ± 5.76
& 88.93 ± 1.88
& 93.40 ± 2.35& 93.39 ± 2.52& \underline{95.08 ± 2.20}  &94.60 ± 2.06
& \textbf{98.46 ± 0.93}
 &$\bullet$97.71 ± 1.63
\\
13 & 21&184
& 96.42 ± 2.37
 & 98.86 ± 1.42
& 99.13 ± 1.01
& \underline{99.89 ± 0.22}
& 96.64 ± 3.30& 98.44 ± 2.49& {99.40 ± 1.03}  &98.63 ± 1.18
& \textbf{100.00 ± 0.00}
 &$\bullet$99.89 ± 0.22
\\
14 & 127&1138
& 93.00 ± 1.85
 & 96.62 ± 2.11
& 96.40 ± 1.43
& 97.35 ± 1.29
& 98.39 ± 0.94& 99.19 ± 0.58& \underline{99.40 ± 0.39}  &99.39 ± 0.59
& \textbf{99.95 ± 0.12}
 &$\bullet$99.93 ± 0.10
\\
15 & 39&347
& 52.80 ± 6.37
 & 77.30 ± 7.36
& 68.88 ± 19.29
& 79.77 ± 10.11
& \underline{98.85 ± 1.79} & 98.49 ± 1.15& 98.71 ± 1.56 &96.72 ± 2.53
& \textbf{99.60 ± 0.26}
 &$\bullet$99.08 ± 0.74
\\
16 & 10&83
& 89.58 ± 4.46
 & 87.62 ± 7.57
& \textbf{99.28 ± 1.45}
& \underline{98.31 ± 2.81}
& {91.02 ± 6.21} & 90.00 ± 6.26& {98.19 ± 3.14} &{96.27 ± 4.69}
& 90.36 ± 2.69
 &$\circ$94.70 ± 3.54
\\\hline
\multicolumn{3}{c|}{OA}& 80.54 ± 0.83
 & 91.10 ± 0.90
& 90.86 ± 1.64
& 93.57 ± 0.42
& 94.99 ± 0.50& 96.94 ± 0.40& {97.50 ± 0.34}  &\underline{97.62 ± 0.32}
& \textbf{98.81 ± 0.21}
 &$\bullet$98.99 ± 0.32
\\
\multicolumn{3}{c|}{AA}& 76.28 ± 17.56
 & 80.64 ± 3.01
& 83.41 ± 2.78
& 77.16 ± 1.52
& 90.42 ± 11.32& 92.56 ± 9.12& {95.36 ± 7.58}  &\textbf{96.64 ± 2.24}
& \underline{96.19 ± 5.20}
 &$\bullet$94.17 ± 2.18
\\
\multicolumn{3}{c|}{$\kappa$}& 77.80 ± 0.93
 & 89.83 ± 1.04
& 89.55 ± 1.88
& 92.66 ± 0.49
& 94.28 ± 0.58& 96.51 ± 0.46& {97.14 ± 0.39}  &\underline{97.28 ± 0.37}
& \textbf{98.64 ± 0.24}
 &$\bullet$98.85 ± 0.37
\\\hline
 \multicolumn{3}{c|}{Time (s)}& 33
  & 262& 302& 127 
 & 659& 570& 621& 59
 & 1102
&-\\
 \hline\hline
\end{tabular}
}
\label{compare1}
\end{table*}

\begin{table*}[ht]
\centering
\caption{Classification Performance of Comparison Methods on \textit{Salinas} with 1.5\% Training Samples. Optimal Values Are Denoted in Bold, and the Second-Best Values Are Underlined. $\bullet$/$\circ$ Indicates Whether the Performance of DSNet on Data Processed by Our Method Is Superior/Inferior to That on the Original Data}
\scalebox{0.75}{
\begin{tabular}{c|cc|c|ccc|cccc|c:c} 
\hline\hline
\multicolumn{4}{c|}{} &\multicolumn{3}{c|}{Deep learning methods}&\multicolumn{4}{c|}{Tensor decomposition methods} &\multicolumn{2}{c}{Ours}\\  \hline 
\#  & Train&Test & Origin    & SpeFormer \citep{SpectralFormer} &  DSNet \citep{DSCNet} &HyperDID \citep{HyperDID}   & LSSTRPCA \citep{2020LSSTRPCA}    & $\mathrm{S^3LRR}$ \citep{2018S3LRR}        & LPGTRPCA \citep{2022LPGTRPCA} &TensorSSA \citep{2023TensorSSA} & Proposed               &DSNet+Ours\\\hline
1 & 31&1978
& 98.69 ± 1.34& 93.33 ± 9.55& 95.28 ± 5.62&  97.14 ± 2.93& 97.89 ± 0.77& 96.36 ± 3.30& 98.88 ± 0.40&\underline{99.24 ± 1.23}& \textbf{99.86 ± 0.24}&$\bullet$99.97 ± 0.06
\\
2 & 56&3670
& 99.40 ± 0.47& 98.85 ± 0.75& \textbf{99.97 ± 0.20}&  99.18 ± 1.13& 98.32 ± 0.92& 97.87 ± 1.97& 98.58 ± 1.16&99.34 ± 0.83& \underline{99.95 ± 0.08}&$\circ$99.91 ± 0.19
\\
3 & 30&1946
& 98.00 ± 1.57& 79.01 ± 16.06& \underline{98.81 ± 1.31}&  98.61 ± 0.68& 94.55 ± 3.22& 96.61 ± 2.80& 95.70 ± 4.74&98.84 ± 0.49& \textbf{99.87 ± 0.25}&$\bullet$99.98 ± 0.04
\\
4 & 21&1373
& \textbf{99.40 ± 0.29}& 98.30 ± 0.67& 98.79 ± 1.52&  \textbf{99.40 ± 0.52}& 97.92 ± 1.07& 92.37 ± 4.01& 94.80 ± 5.07&\underline{98.99 ± 0.29}& 96.37 ± 1.08&$\circ$92.88 ± 7.86
\\
5 & 41&2637
& 96.87 ± 1.40& 98.19 ± 1.51& \textbf{99.64 ± 0.29}&  \underline{98.74 ± 1.25}& 95.22 ± 2.12& 95.06 ± 2.84& 96.36 ± 1.30&96.65 ± 1.47& 98.23 ± 0.52&$\circ$98.46 ± 0.95
\\
6 & 60&3899
& 99.38 ± 0.42& 99.79 ± 0.37& \textbf{100.00 ± 0.00}&  99.93 ± 0.13& 99.84 ± 0.14& 98.96 ± 0.83& 99.63 ± 0.30&99.87 ± 0.09& \underline{99.94 ± 0.01}&$\circ$99.90 ± 0.14
\\
7 & 54&3525
& \underline{99.52 ± 0.16}& 97.49 ± 1.55& 99.43 ± 0.18&  99.41 ± 0.38& 98.59 ± 1.06& 98.13 ± 0.93& 97.65 ± 0.85&99.60 ± 0.26& \textbf{99.80 ± 0.06}&$\bullet$99.81 ± 0.26
\\
8 & 170&11101
& 82.50 ± 1.19& 83.07 ± 8.43& 89.71 ± 4.53&  90.75 ± 2.90& \textbf{98.50 ± 0.44}&\underline{98.23 ± 0.38}& 97.55 ± 1.20&95.00 ± 0.68& 98.22 ± 1.27&$\bullet$99.55 ± 0.18
\\
9 & 94&6109
& 99.09 ± 0.76& 98.77 ± 0.59& 99.68 ± 0.52&  99.74 ± 0.26& 98.90 ± 0.84& 98.35 ± 1.45& \underline{99.77 ± 0.25}&99.50 ± 0.43& \textbf{99.87 ± 0.14}&$\bullet$99.76 ± 0.32
\\
10 & 50&3228
& 93.18 ± 1.28& 92.03 ± 3.08& \underline{96.38 ± 1.05}&  94.59 ± 1.22& 94.31 ± 1.42& 95.57 ± 0.96& \textbf{97.68 ± 1.07}&95.70 ± 1.60& 95.84 ± 1.11&$\bullet$98.46 ± 1.82
\\
11 & 17&1051
& 93.53 ± 4.81& 88.54 ± 5.19& \underline{97.37 ± 1.73}&  94.39 ± 5.08& 94.23 ± 2.54& 89.46 ± 8.53& 91.95 ± 2.74&97.94 ± 1.93& \textbf{98.71 ± 1.95}&$\circ$96.82 ± 5.66
\\
12 & 29&1898
& 98.89 ± 1.69& 98.56 ± 2.29& \textbf{100.00 ± 0.00}&  98.68 ± 2.63& 98.96 ± 0.69& 95.34 ± 6.37& 96.72 ± 0.48&99.97 ± 0.05& \underline{99.60 ± 0.41}&$\circ$99.40 ± 0.84
\\
13 & 14&902
& 98.12 ± 0.92& 98.43 ± 1.81& \textbf{99.87 ± 0.21}&  \underline{99.84 ± 0.09}& 94.30 ± 3.38& 84.17 ± 10.19& 91.82 ± 1.96&97.98 ± 1.50& 97.69 ± 1.42&$\circ$94.41 ± 5.46
\\
14 & 17&1053
& 91.02 ± 2.19& 95.21 ± 2.03& \textbf{98.25 ± 0.70}&  94.85 ± 1.51& 93.66 ± 2.67& 93.90 ± 2.64& 93.62 ± 8.35&95.78 ± 2.69& \underline{97.17 ± 0.87}&$\circ$98.12 ± 1.79
\\
15 & 110&7158
& 71.08 ± 3.27& 75.95 ± 7.91& 72.94 ± 12.49&  75.14 ± 3.97& 97.72 ± 0.62& \underline{98.29 ± 0.90}& 96.87 ± 1.35&96.28 ± 1.66& \textbf{98.61 ± 0.46}&$\bullet$99.42 ± 0.21
\\
16 & 28&1779
& 97.31 ± 0.94& 92.46 ± 2.91& \underline{98.85 ± 0.62}&  97.64 ± 1.36& 97.97 ± 1.12& 97.07 ± 3.58& \textbf{99.07 ± 0.54}&98.23 ± 0.49& 98.23 ± 0.16&$\bullet$99.33 ± 0.66
\\\hline
\multicolumn{3}{c|}{}& 91.08 ± 0.43& 90.56 ± 1.61& 93.53 ± 1.23&  93.70 ± 0.48& \underline{97.68 ± 0.28}& 97.04 ± 0.47& 97.54 ± 0.28&97.58 ± 0.33& \textbf{98.77 ± 0.32}&$\bullet$99.18 ± 0.16
\\
\multicolumn{3}{c|}{AA}& 94.75 ± 7.75& 93.00 ± 1.81& 96.56 ± 0.76&  96.13 ± 0.29& 96.93 ± 2.12& 95.36 ± 3.91& 96.66 ± 2.52&\underline{98.06 ± 1.66}& \textbf{98.62 ± 1.33}&$\bullet$98.51 ± 0.37
\\
\multicolumn{3}{c|}{$\kappa$}& 90.06 ± 0.48& 89.50 ± 1.79& 92.79 ± 1.39&  92.98 ± 0.53& \underline{97.42 ± 0.32}& 96.70 ± 0.53& 97.26 ± 0.32&97.31 ± 0.37& \textbf{98.63 ± 0.35}&$\bullet$99.09 ± 0.18
\\\hline
 \multicolumn{3}{c|}{Time (s)}& 39& 1281& 898& 271& 1452& 1892& 1506& 314
& 6015
&-\\
\hline\hline
\end{tabular}
}
\label{compare2}
\end{table*}

\begin{table*}[ht]
\centering
\caption{Classification Performance of Comparison Methods on \textit{Pavia University} with 0.5\% Training Samples. Optimal Values Are Denoted in Bold, and the Second-Best Values Are Underlined. $\bullet$/$\circ$ Indicates Whether the Performance of DSNet on Data Processed by Our Method Is Superior/Inferior to That on the Original Data}
\scalebox{0.75}{
\begin{tabular}{c|cc|c|ccc|cccc|c:c} 
\hline\hline
\multicolumn{4}{c|}{} &\multicolumn{3}{c|}{Deep learning methods}&\multicolumn{4}{c|}{Tensor decomposition methods} &\multicolumn{2}{c}{Ours}\\  \hline 
\#  & Train&Test & Origin    & SpeFormer \citep{SpectralFormer} &  DSNet \citep{DSCNet} &HyperDID \citep{HyperDID}   & LSSTRPCA \citep{2020LSSTRPCA}    & $\mathrm{S^3LRR}$ \citep{2018S3LRR}        & LPGTRPCA \citep{2022LPGTRPCA} &TensorSSA \citep{2023TensorSSA} & Proposed               &DSNet+Ours\\\hline
1 & 34&6597
& 85.66 ± 3.22
 & 73.15 ± 6.86
& \underline{90.64 ± 5.17}
& \textbf{94.00 ± 0.88}& 85.34 ± 3.35& {88.61 ± 2.98}& 85.27 ± 3.50 &{90.97 ± 3.08} 
& {86.65 ± 2.90}
 &$\bullet$92.25 ± 7.13
\\
2 & 94&18555
& 93.90 ± 2.40
 & 95.77 ± 3.57
& \textbf{99.72 ± 0.14}
& 98.08 ± 0.90& {96.52 ± 1.13} & 95.07 ± 1.80& 95.31 ± 1.99 &{96.92 ± 1.62} 
& \underline{99.25 ± 0.55}
 &$\circ$99.24 ± 0.96
\\
3 & 11&2088
& 63.30 ± 10.42
 & 48.52 ± 11.24
& 66.88 ± 20.78
& 43.74 ± 12.03& 64.03 ± 8.24& 63.24 ± 9.24& {67.87 ± 5.73}  &\textbf{82.31 ± 6.98} 
& \underline{75.72 ± 7.40}
 &$\circ$62.80 ± 17.66
\\
4 & 16&3048
& 81.76 ± 5.69
 & 85.41 ± 4.41
& 89.97 ± 3.28
&  \underline{90.77 ± 5.46}&{90.48 ± 4.11} & 82.06 ± 8.29& 86.98 ± 4.95 &87.05 ± 3.55 
& \textbf{90.87 ± 2.64}
 &$\bullet$93.76 ± 1.29
\\
5 & 7&1338
& 94.79 ± 10.24
 & \textbf{99.40 ± 0.75}
& 96.82 ± 3.52
& \underline{98.89 ± 1.25}& {98.65 ± 0.99} & {99.19 ± 0.72}& 97.88 ± 3.42 &95.73 ± 6.97 
& 97.94 ± 4.70
 &$\bullet$99.48 ± 0.43
\\
6 & 26&5003
& 74.12 ± 5.43
 & 42.06 ± 12.26
& 84.27 ± 4.34
& 84.22 ± 3.57& 80.02 ± 5.99& {85.74 ± 6.67} & 74.61 ± 8.31 &\underline{85.76 ± 4.39} 
& \textbf{95.26 ± 2.63}
 &$\bullet$98.14 ± 1.33
\\
7 & 7&1323
& 68.70 ± 10.47
 & 62.12 ± 6.83
& 78.82 ± 17.91
& 70.04 ± 3.58& 74.03 ± 8.77& {80.90 ± 10.05} & 72.06 ± 10.17 &\textbf{88.51 ± 6.27} 
& \underline{82.44 ± 8.04}
 &$\bullet$82.46 ± 13.68
\\
8 & 19&3663
& 77.05 ± 6.35
 & 74.48 ± 8.66
& \textbf{90.37 ± 4.69}
& 84.64 ± 8.10& 78.41 ± 5.83& 74.77 ± 7.08& {80.21 ± 5.67}  &{83.60 ± 4.82} 
& \underline{87.74 ± 3.87}
 &$\bullet$93.75 ± 6.10
\\
9 & 5&942
& \underline{99.72 ± 0.13} 
 & 92.21 ± 3.53
& 96.82 ± 2.65
& 99.15 ± 0.97& 99.68 ± 0.15& 99.50 ± 0.29& \textbf{99.76 ± 0.13} &98.63 ± 0.59 
& 99.70 ± 0.27
 &$\bullet$99.00 ± 1.16
\\\hline
\multicolumn{3}{c|}{OA}& 85.85 ± 1.41
 & 80.05 ± 1.69
& \underline{92.58 ± 1.33}
& 90.65 ± 0.47& {88.70 ± 1.20} & 88.52 ± 1.51& 87.53 ± 1.15 &{91.85 ± 1.24} 
& \textbf{93.53 ± 0.73}
 &$\bullet$94.85 ± 1.73
\\
\multicolumn{3}{c|}{AA}& 82.11 ± 12.48
 & 74.79 ± 1.86
& 88.26 ± 2.20
& 84.84 ± 0.78& 85.24 ± 12.21& {85.45 ± 11.88} & 84.44 ± 11.63 &\underline{89.94 ± 5.97} 
& \textbf{90.62 ± 8.26}
 &$\bullet$91.21 ± 3.59
\\
\multicolumn{3}{c|}{$\kappa$}& 81.12 ± 1.83
 & 72.95 ± 2.36
& \underline{90.07 ± 1.78}
& 87.50 ± 0.59& {84.95 ± 1.61} & 84.72 ± 2.01& 83.36 ± 1.55 &{89.15 ± 1.63} 
& \textbf{91.40 ± 0.97}
 &$\bullet$93.17 ± 2.30
\\\hline
 \multicolumn{3}{c|}{Time (s)}& 11& 969& 697& 246& 1895& 2449& 1392& 324 
& 3751
& - \\
 \hline\hline
\end{tabular}
}
\label{compare3}
\end{table*}

\begin{table*}[ht]
\centering
\caption{Classification Performance of Comparison Methods on \textit{WHU-Hi-LongKou} with 1\% Training Samples. Optimal Values Are Denoted in Bold, and the Second-Best Values Are Underlined. $\bullet$/$\circ$ Indicates Whether the Performance of DSNet on Data Processed by Our Method Is Superior/Inferior to That on the Original Data}
\scalebox{0.75}{
\begin{tabular}{c|cc|c|ccc|cccc|c:c} 
\hline\hline
\multicolumn{4}{c|}{} &\multicolumn{3}{c|}{Deep learning methods}&\multicolumn{4}{c|}{Tensor decomposition methods} &\multicolumn{2}{c}{Ours}\\  \hline 
\#  & Train&Test & Origin    & SpeFormer \citep{SpectralFormer} &  DSNet \citep{DSCNet} &HyperDID \citep{HyperDID}   & LSSTRPCA \citep{2020LSSTRPCA}    & $\mathrm{S^3LRR}$ \citep{2018S3LRR}        & LPGTRPCA \citep{2022LPGTRPCA} &TensorSSA \citep{2023TensorSSA} & Proposed               &DSNet+Ours\\\hline
1 & 15&1396
& 88.38 ± 4.13
 & 92.39 ± 2.60
& 96.88 ± 1.70
& \underline{97.03 ± 1.58}& 91.74 ± 5.37& 92.58 ± 3.81& {94.96 ± 3.27}  &{96.78 ± 1.54} 
& \textbf{99.08 ± 0.75}
 &$\bullet$99.70 ± 0.53
\\
2 & 36&3532
& 87.88 ± 3.32
 & 88.78 ± 3.57
& \underline{96.70 ± 1.18}
& 96.19 ± 2.73& {95.29 ± 2.05} & 91.94 ± 2.67& 94.61 ± 2.02 &{95.93 ± 2.67} 
& \textbf{98.09 ± 1.49}
 &$\bullet$99.39 ± 0.28
\\
3 & 13&1216
& 73.27 ± 6.44
 & 68.87 ± 23.91
& \underline{96.38 ± 1.58}
& 55.44 ± 29.45& 87.20 ± 5.42& 84.90 ± 4.43& {91.53 ± 4.23}  &\textbf{97.40 ± 3.12} 
& {95.16 ± 1.75}
 &$\bullet$98.95 ± 1.30
\\
4 & 59&5773
& 85.51 ± 2.83
 & 66.73 ± 8.56
& 88.79 ± 3.38
& 90.80 ± 4.29& {91.09 ± 2.40} & 89.62 ± 2.32& 91.09 ± 2.61 &\textbf{96.37 ± 1.43} 
& \underline{95.86 ± 1.79}
 &$\bullet$96.55 ± 1.67
\\
5 & 14&1324
& 71.11 ± 5.00
 & 74.15 ± 5.14
& 77.61 ± 4.17
& 48.69 ± 30.34& 73.98 ± 5.01& {76.98 ± 6.18} & 69.09 ± 8.42 &\underline{90.74 ± 3.90} 
& \textbf{92.64 ± 2.82}
 &$\bullet$95.47 ± 3.19
\\
6 & 7&665
& 87.43 ± 7.51
 & 93.80 ± 5.88
& \underline{99.28 ± 0.93}
& 97.02 ± 5.08& 95.85 ± 4.51& 93.11 ± 6.46& {96.28 ± 3.57}  &93.64 ± 5.41 
& \textbf{99.98 ± 0.07}
 &$\bullet$99.97 ± 0.06
\\
7 & 207&20403
& \underline{99.98 ± 0.01} 
 & 99.95 ± 0.07
& 99.83 ± 0.06
& \textbf{99.99 ± 0.01}& 99.95 ± 0.04& \textbf{99.99 ± 0.01}& 99.88 ± 0.13 &99.87 ± 0.08 
& 99.94 ± 0.04
 &$\circ$99.71 ± 0.17
\\
8 & 25&2428
& 79.61 ± 6.65
 & 80.71 ± 10.24
& 88.36 ± 4.55
& \textbf{90.74 ± 6.02}& 80.36 ± 4.41& 84.19 ± 4.26& {84.88 ± 6.23}  &84.62 ± 5.82
& \underline{90.23 ± 3.39}
 &$\bullet$93.52 ± 6.10
\\
9 & 15&1435
& 55.93 ± 5.65
 & \textbf{80.47 ± 6.92}
& 75.16 ± 6.13
& 43.48 ± 21.76& 54.32 ± 6.80& 58.43 ± 7.29& {61.91 ± 5.46}  &{77.07 ± 8.72} 
& \underline{80.33 ± 8.28}
 &$\bullet$84.68 ± 6.56
\\\hline
\multicolumn{3}{c|}{OA}& 91.23 ± 0.66
 & 89.66 ± 1.26
& 95.21 ± 0.43
& 92.18 ± 1.38& 93.54 ± 0.58& 93.44 ± 0.64& {94.11 ± 0.56}  &\underline{96.53 ± 0.55} 
& \textbf{97.36 ± 0.46}
 &$\bullet$98.07 ± 0.29
\\
\multicolumn{3}{c|}{AA}& 81.01 ± 12.81
 & 82.87 ± 3.32
& 91.00 ± 0.69
& 79.93 ± 4.79& 85.53 ± 14.21& 85.75 ± 12.16& {87.14 ± 13.06}  &\underline{92.49 ± 7.31}
& \textbf{94.59 ± 6.31}
 &$\bullet$96.44 ± 0.54
\\
\multicolumn{3}{c|}{$\kappa$}& 86.96 ± 0.97
 & 84.72 ± 1.86
& 92.90 ± 0.63
& 88.31 ± 2.08& 90.39 ± 0.86& 90.25 ± 0.95& {91.24 ± 0.83}  &\underline{94.85 ± 0.81} 
& \textbf{96.08 ± 0.68}
 &$\bullet$97.14 ± 0.43
\\\hline
 \multicolumn{3}{c|}{Time (s)}& 15& 882& 669& 199& 791& 1628& 815& 127 
& 1311
&-\\
 \hline\hline
\end{tabular}
}
\label{compare4}
\end{table*}

\subsection{Experiment Setup and Compared Methods}

To evaluate the representation ability, we performed the classification task on the learned representation. Specifically, a typical SVM classifier equipped with an RBF kernel was implemented as the classifier, and three metrics, including the overall classification accuracy (OA), the average class classification accuracy (AA), and the kappa coefficient ($\kappa$), were used to measure the classification accuracy. We repeated the classifier 5 times to obtain the average of OA, AA, and $\kappa$ across all compared methods.

We compared the classification performance of our method with seven state-of-the-art methods, including four tensor-based methods, i.e., LSSTRPCA \citep{2020LSSTRPCA}, $\mathrm{S^3LRR}$ \citep{2018S3LRR}, LPGTRPCA \citep{2022LPGTRPCA}, TensorSSA \citep{2023TensorSSA} and three deep learning methods, i.e., SpeFormer \citep{SpectralFormer}, DSNet \citep{DSCNet}, HyperDID \citep{HyperDID}. For a fair comparison, we used the codes provided by their inventors, and the parameter settings of these methods were tuned to optimal ones. Table \ref{parameters_opt} presents the parameter settings for the proposed method. The details of the compared methods are listed as follows:

\begin{enumerate}
\item{{LSSTRPCA \citep{2020LSSTRPCA}}: is a lateral-slice sparse tensor RPCA with a tensor $l_{2,1}$ norm for sparse component to gross errors or outliers.}

\item{{$\mathrm{S^3LRR}$ \citep{2018S3LRR}}: is a model exploring the low-rank property spectral and spatial domain simultaneously in the 2-D matrix domain.}

\item{{LPGTRPCA \citep{2022LPGTRPCA}}: is a tensor RPCA model with a locality-preserving graph and frontal slice sparsity.}

\item{{TensorSSA \citep{2023TensorSSA}}: is a method of 3D tensor Singular Spectrum Analysis (SSA) that extracts the low-rank features of HSI by decomposing the trajectory tensor and reconstructing it with low-rank approximation.}

\item{{SpeFormer \citep{SpectralFormer}}: is a transformer-based model for the HSI classification task, leveraging group-wise spectral embeddings and cross-layer adaptive fusion to improve the capture of spectral and spatial features.}

\item{{DSNet \citep{DSCNet}}: is a dual-branch subpixel-guided network that integrates a deep autoencoder unmixing architecture to enhance hyperspectral image classification by fusing subpixel and pixel-level features.}

\item{{HyperDID \citep{HyperDID}}: is a hyperspectral intrinsic image decomposition framework incorporating a deep feature embedding to separate environment-related and category-related features for improved hyperspectral image classification.}

\end{enumerate}

All the tensor-based methods were conducted on a Windows 10 server equipped with two Intel Xeon Gold 6248R CPUs. The deep-learning methods were carried out on a Linux server with an AMD EPYC 7642 CPU and an NVIDIA RTX 4090 GPU.

\begin{figure*}[ht]
\centering
 \begin{minipage}{4cm}
 \centering
  \includegraphics[width=1.1\textwidth]{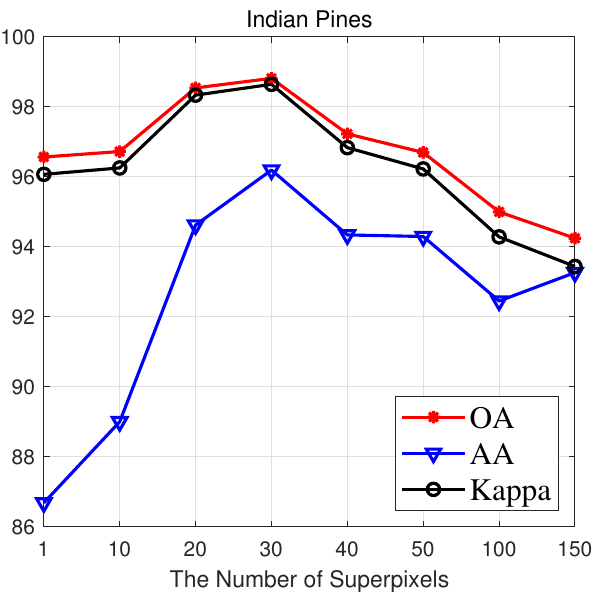}
  \end{minipage}\hspace{5mm}
  \begin{minipage}{4cm}
  \centering
  \includegraphics[width=1.1\textwidth]{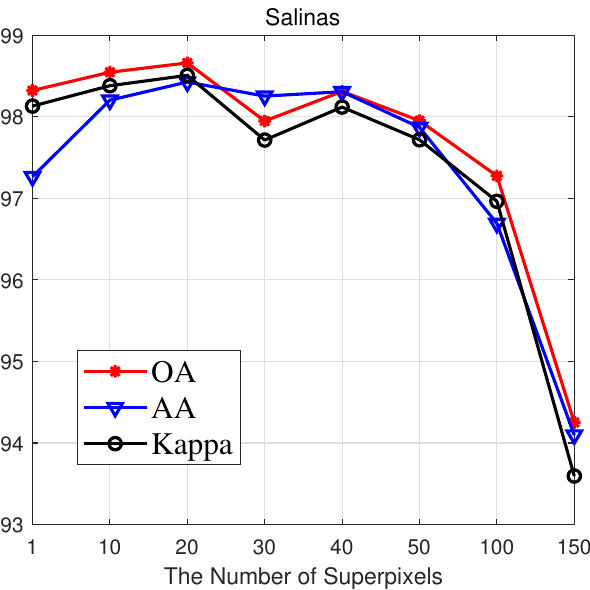}
    \end{minipage}\hspace{5mm}
  \begin{minipage}{4cm}
  \centering
  \includegraphics[width=1.1\textwidth]{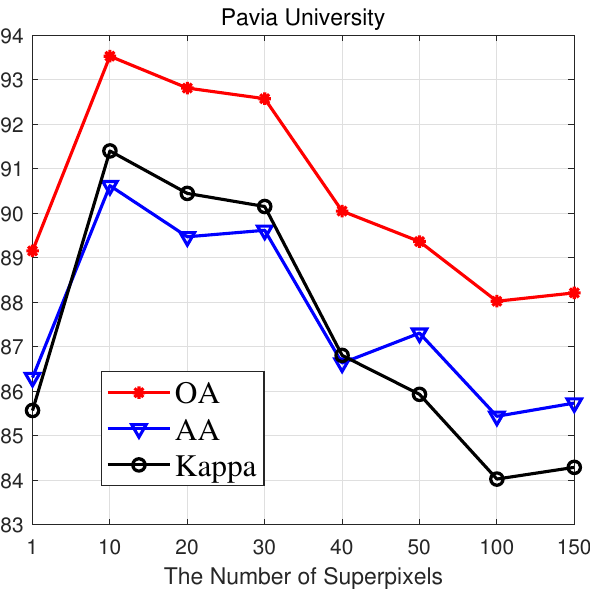}
    \end{minipage}\hspace{5mm}
  \begin{minipage}{4cm}
  \centering
  \includegraphics[width=1.1\textwidth]{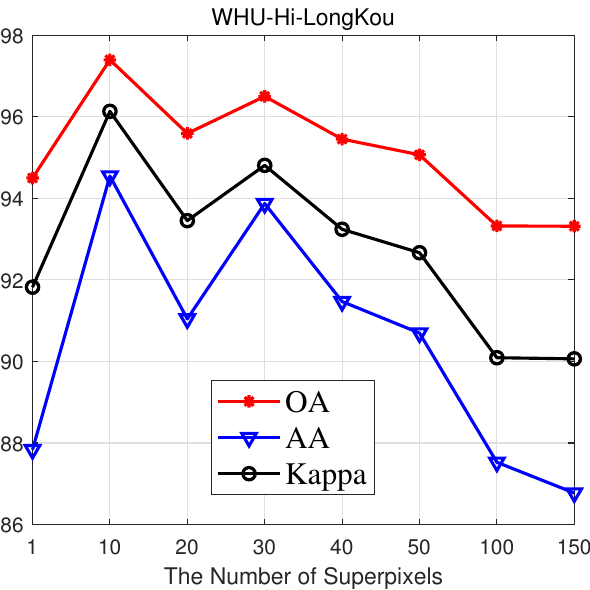}
  \end{minipage}
\caption{Illustration of the influence of the number of superpixels of our model on classification performance on four datasets.}
  \label{number_of_superpixle}
\end{figure*}

\begin{figure*}[ht]
\centering
 \begin{minipage}{4cm}
 \centering
  \includegraphics[width=1.1\textwidth]{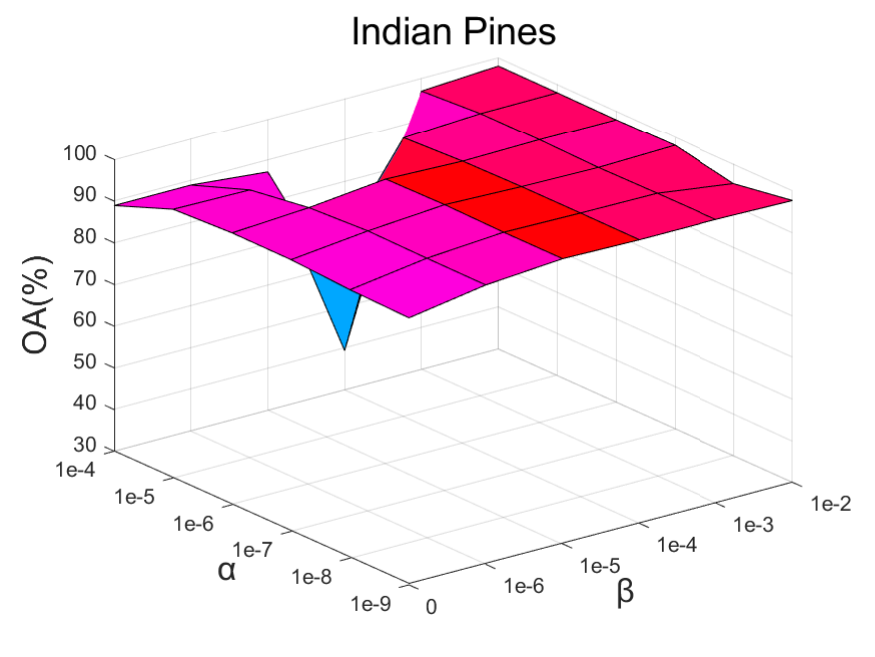} 
  \end{minipage}\hspace{5mm}
  \begin{minipage}{4cm}
  \centering
  \includegraphics[width=1.1\textwidth]{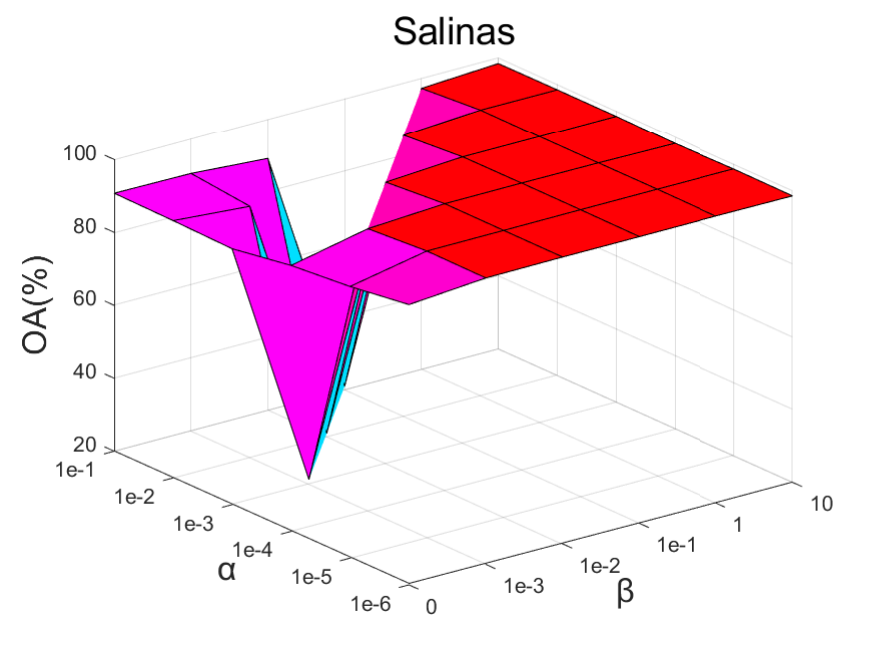}
    \end{minipage}\hspace{5mm}
  \begin{minipage}{4cm}
  \centering
  \includegraphics[width=1.1\textwidth]{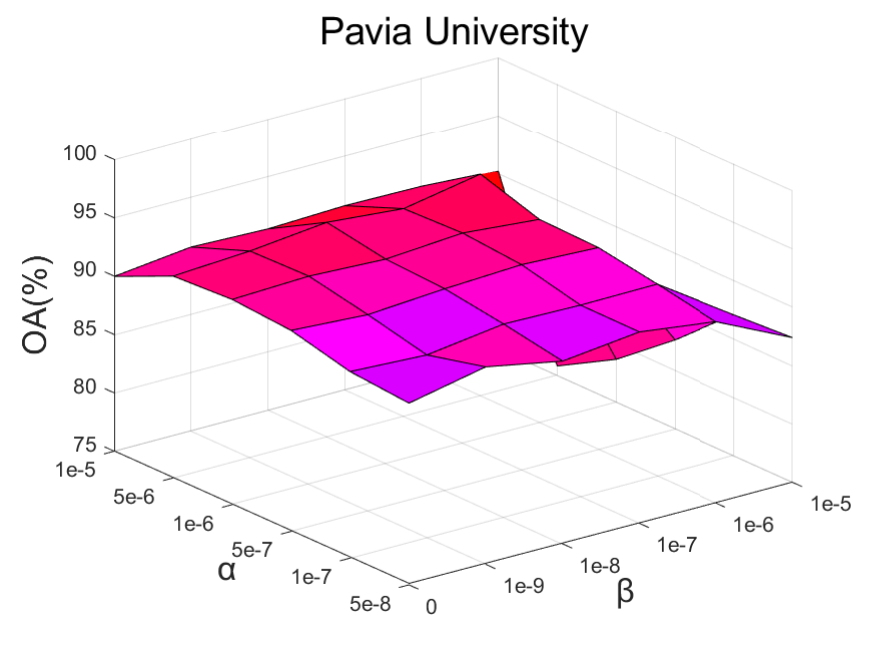}  
    \end{minipage}\hspace{5mm}
  \begin{minipage}{4cm}
  \centering
  \includegraphics[width=1.1\textwidth]{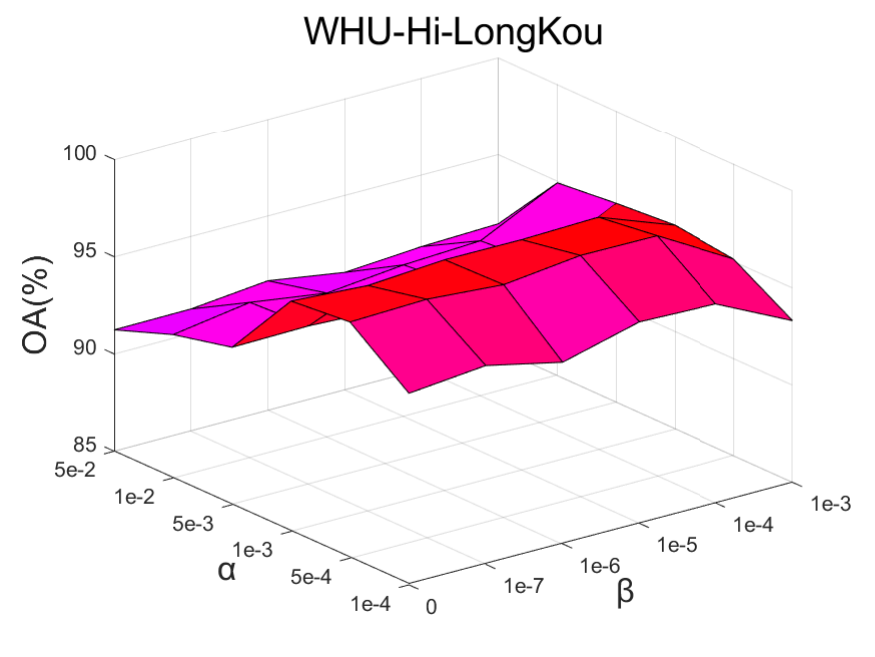}
  \end{minipage}
\caption{Influence of the two hyper-parameters ($\alpha$ and $\beta$) of our model with respect to OA (\%) on four datasets.}
  \label{two_parameters}
\end{figure*}

\subsection{Comparison with State-of-the-Art Methods}

As shown in Fig. \ref{compare-all}, the proposed method performs better than other methods under various training percentages on all datasets. Especially, when the training percentage is relatively small, the superiority of our method becomes particularly evident. For example, when the training percentage is 0.2\% on \textit{WHU-Hi-LongKou}, our method is 5\% higher than other the best compared method in terms of OA.

Table \ref{compare1} lists the classification accuracy on all classes of all methods in comparison on \textit{Indian Pines}, in which the training percentage is 10\%. According to the results, we can see that our proposed method produces the optimal OA (98.81\%) and $\kappa$ (98.64\%) while the second best OA and $\kappa$ are only 97.62\% and 97.28\%. Additionally, our proposed method achieves a suboptimal performance in AA, slightly lower than TensorSSA. For class-specific accuracy, the proposed method achieves the best performance on most classes except for classes 3, 4, 7, 9 and 16. For the classes with small number of pixels, i.e., classes 1, 5, 12, our method improves the accuracy significantly.

Table \ref{compare2} shows the classification accuracy on all classes of all methods in comparison on \textit{Salinas}, where the training percentage is 1.5\%. It can be observed that our proposed method achieves the best performance in OA, AA and $\kappa$. Specifically, our method produced the highest values in most classes, especially for classes with limited pixels, i.e., classes 1 and 11.

Similar results can be observed on \textit{Pavia University} and \textit{WHU-Hi-LongKou} as shown in Tables \ref{compare3}-\ref{compare4}. Especially, on \textit{WHU-Hi-LongKou}, our method achieves the optimal and suboptimal accuracy on 7 out of 9 classes, with the exception of classes 3 and 7.

{As a comparison, the performance of deep learning based methods is limited, which can be attributed to the small number of training samples and the insufficient discriminative power of the original data. Our method is designed to learn discriminative low-rank representations, which not only improve the performance of SVM-based classification but also significantly enhance deep learning based approaches. As demonstrated in the last column of Tables \ref{compare1}-\ref{compare4}, we applied DSNet \citep{DSCNet} to classify the data processed by our method. The results show a substantial improvement in classification performance compared to the original DSNet method. Specifically, on \textit{Indian Pines}, \textit{Pavia University}, and \textit{WHU-Hi-LongKou}, classification accuracy increased in 15 out of 16 classes, 7 out of 9 classes, and 8 out of 9 classes, respectively. Furthermore, categories with limited training samples showed significant improvements. For example, on \textit{Indian Pines}, the accuracy of class 1 with only 5 training samples increased from 62.44\% to 90.73\%.}

From the classification maps in Fig. \ref{comparefig1}, we can find that our proposed method has less misclassified pixels than others, which further demonstrates the superiority of our method. {The classification maps on \textit{Salinas}, \textit{Pavia University}, and \textit{WHU-Hi-LongKou} can be found in Fig. S3-S5 of the supplementary material.}

{Additionally, we conducted runtime tests to evaluate the time efficiency of our method. As shown in Tables \ref{compare1}-\ref{compare4}, our method achieves runtime durations of 1102, 6015, 3751, and 1311 seconds across four datasets, respectively. Compared with other methods, our approach requires more time on \textit{Indian Pines}, \textit{Salinas}, \textit{Pavia University} and is competitive on \textit{WHU-Hi-LongKou}. Although our runtime is slightly higher, the significant performance improvement outweighs the time cost, demonstrating the trade-off between efficiency and effectiveness.}

\begin{table*}
\centering
\caption{Ablation Study on \textit{Indian Pines} with 10\% Training Samples. Optimal Values Are Denoted in Bold, and the Second-Best Values Are Underlined}
\scalebox{0.8}{
{\begin{tabular}{c|cc|ccccc|ccc|c} 
\hline\hline
\#     & Train&Test & Origin              & RPCA \citep{2011rpca}              &Super-RPCA& TRPCA \citep{2020TRPCA}             &Patch-TRPCA& M1                     & M2                       &M3& Proposed                      \\\hline
1 & 5&41 & 63.75 ± 14.75& 78.38 ± 12.23 &94.63 ± 6.54& 83.00 ± 11.85 &84.00 ± 5.76 & 93.88 ± 6.46 & \textbf{97.75 ± 0.77} & \underline{95.00 ± 0.00} & {94.00 ± 1.37}
\\ 
2 & 143&1285 & 79.19 ± 2.49& 82.90 ± 2.06 &89.06 ± 1.10& 93.75 ± 1.46 &92.40 ± 1.85& 96.14 ± 1.50 & {96.34 ± 1.08}  &\underline{96.36 ± 1.45} & \textbf{98.72 ± 0.77}
\\ 
3 & 83&747 & 71.03 ± 3.02& 77.13 ± 2.68 &89.45 ± 2.43 & 93.12 ± 1.94 &90.62 ± 2.46 & 93.55 ± 2.17 & {94.56 ± 2.01}  &\underline{95.23 ± 1.20} & \textbf{96.86 ± 1.38}
\\ 
4 & 24&213 & 55.00 ± 6.32& 68.54 ± 6.55 &78.59 ± 11.46 & {88.57 ± 5.38}  &89.67 ± 3.94 & 87.18 ± 6.07& 85.89 ± 5.33 &\underline{91.83 ± 5.21} & \textbf{94.46 ± 2.88}
\\ 
5 & 49&434 & 90.18 ± 2.33& 91.64 ± 2.89 &\underline{97.24 ± 1.37}& 95.38 ± 2.65 &92.84 ± 3.25 & {96.10 ± 2.08}  & 95.92 ± 2.28 &{96.54 ± 1.39}& \textbf{98.38 ± 1.02}
\\ 
6 & 73&657 & 95.60 ± 1.50& 96.92 ± 1.31 &99.06 ± 1.19& 99.21 ± 0.50 &99.15 ± 0.91 & 99.01 ± 0.93& \underline{99.34 ± 1.10}   &99.33 ± 0.56 & \textbf{99.54 ± 1.02}
\\ 
7 & 3&25 & 75.80 ± 12.34& 85.20 ± 6.63 &88.80 ± 1.79& 81.40 ± 10.96 &90.40 ± 11.52 & {91.80 ± 9.04}  & \underline{94.80 ± 3.69} &\textbf{96.80 ± 3.35} & 88.00 ± 12.00
\\ 
8 & 48&430 & 96.84 ± 1.81& 98.25 ± 1.03 &\textbf{100.00 ± 0.00}& 99.28 ± 0.79 &97.72 ± 1.42 & \underline{99.98 ± 0.10}  & \textbf{100.00 ± 0.00} &\textbf{100.00 ± 0.00} & \textbf{100.00 ± 0.00}
\\ 
9 & 2&18& 37.22 ± 13.50& 45.00 ± 15.18 &\textbf{100.00 ± 0.00}& 76.39 ± 16.31 &81.11 ± 13.38 & \underline{83.61 ± 20.82} & 82.50 ± 22.10  &61.11 ± 17.57 & 82.22 ± 18.17
\\ 
10 & 98&874 & 74.30 ± 2.69& 76.32 ± 3.30 &90.14 ± 1.48& 91.37 ± 1.90 &92.55 ± 2.75 & 93.01 ± 2.27 & {94.13 ± 1.88}  &\underline{94.54 ± 1.26} & \textbf{99.11 ± 0.83}
\\ 
11 & 246&2209 & 79.41 ± 2.44& 83.45 ± 1.17 &97.83 ± 1.41& 94.70 ± 1.19 &93.66 ± 2.75 & {97.12 ± 0.98}  & 97.06 ± 0.89 &\underline{98.40 ± 0.33} & \textbf{99.38 ± 0.40}
\\ 
12 & 60&533& 70.37 ± 3.26& 81.60 ± 4.09 &88.11 ± 4.03& 90.37 ± 3.93 &85.54 ± 3.45 & 93.66 ± 2.46 & {94.85 ± 2.52}  &\underline{97.36 ± 1.71} & \textbf{98.46 ± 0.93}
\\ 
13 & 21&184 & 96.42 ± 2.37& 98.52 ± 1.74 &99.46 ± 0.00& 98.88 ± 1.03 &99.45 ± 0.77 & {99.34 ± 0.38}  & 99.32 ± 0.39 &\underline{99.78 ± 0.49} & \textbf{100.00 ± 0.00}
\\ 
14 & 127&1138 & 93.00 ± 1.85& 94.55 ± 1.71 &99.75 ± 0.11& 99.03 ± 0.40 &97.74 ± 1.26 & 99.37 ± 0.33 & {99.66 ± 0.22}  &\underline{99.82 ± 0.09} & \textbf{99.95 ± 0.12}
\\ 
15 & 39&347 & 52.80 ± 6.37& 58.48 ± 6.78 &96.60 ± 1.67& {98.94 ± 0.84}  &96.32 ± 1.72 & 92.82 ± 4.20& 94.70 ± 2.72 &\underline{99.08 ± 0.74} & \textbf{99.60 ± 0.26}
\\ 
16 & 10&83 & 89.58 ± 4.46& 90.96 ± 4.16 &90.60 ± 5.14& 87.95 ± 8.26 &75.66 ± 11.12 & \underline{94.88 ± 3.31}  & \textbf{95.06 ± 3.43} &94.22 ± 4.92 & 90.36 ± 2.69
\\\hline
\multicolumn{3}{c|}{OA} & 80.54 ± 0.83& 84.46 ± 0.61 &94.48 ± 0.43& 94.89 ± 0.61 &93.59 ± 0.51 & 96.18 ± 0.52 & {96.57 ± 0.29}  &\underline{97.43 ± 0.27} & \textbf{98.81 ± 0.21}
\\ 
\multicolumn{3}{c|}{AA} & 76.28 ± 17.56& 81.74 ± 14.78 &93.71 ± 6.12& 91.96 ± 7.03 &91.18 ± 6.76 & 94.46 ± 4.44 & \underline{95.12 ± 4.75}  &94.71 ± 9.27 & \textbf{96.19 ± 5.20}
\\ 
\multicolumn{3}{c|}{$\kappa$} & 77.80 ± 0.93& 82.28 ± 0.69 &93.69 ± 0.49& 94.18 ± 0.69 &92.70 ± 0.58 & 95.64 ± 0.60 & {96.08 ± 0.34}  &\underline{97.07 ± 0.31} & \textbf{98.64 ± 0.24}
\\
\hline\hline
\end{tabular}}}
\label{study1}
\end{table*}

\subsection{Hype-Parameter}

In this section, we investigated how the three hyper-parameters, i.e. $\lambda$, $\beta$ and the number of superpixels $n$, affect the performance of the proposed ITLRR. For each superpixel block \(\mathcal{X}_i \), the regularization parameter \(\lambda_i\) is related to the size of the block, i.e., \(\lambda_i = \alpha / \sqrt{\max(w_i, b_i)n_3}\). Therefore, we actually studied the effect of $\alpha$ instead of $\lambda_i$.

\textit{1) Influence of the Number of Superpixels:} Fig. \ref{number_of_superpixle} shows the classification performance of four datasets under different superpixel numbers. According to Fig. \ref{number_of_superpixle}, the performance of our model on all the datasets increases with the number of superpixels from 1 to more, especially AA improves significantly. This proves that the introduction of superpixel segmentation in our method is helpful. Besides, too many superpixels will degrade the performance. That is because too many superpixels will result in pixel points in homogeneous areas being divided into different superpixels, decreasing the similarity of pixel points in homogeneous areas. From the results, we can obtain that the optimal number of superpixels for the four datasets are 30, 20, 10 and 10, respectively.

\textit{2) Influence of Parameter $\alpha$:} The OA of the proposed ITLRR with different parameters $\alpha$ and $\beta$ is demonstrated in Fig. \ref{two_parameters}. The numbers of superpixels for the datasets were fixed at the optimal ones. For the parameter $\alpha$, our model always produces a high OA on all four datasets with a wide range of $\alpha$. It can be observed that the optimal $\alpha$ with a fixed value of $\beta$ for four datasets are in the intervals [1e-7, 1e-9], [1e-4, 1e-6], [1e-5, 1e-6] and [1e-3, 1e-4] respectively, which demonstrate the robustness with respect to $\alpha$. Generally, the value of $\alpha$ depends on the severity of spectral variations, i.e., the more serious, the smaller. Obviously, the optimal value of $\alpha$ for \textit{WHU-Hi-LongKou} is greater than that of other three datasets. That is because that other three datasets suffered from more serious spectral variations than acquired \textit{WHU-Hi-LongKou}, and a smaller $\alpha$ enables the model to remove more noises. 

\textit{3) Influence of Parameter $\beta$:} On all four datasets, the optimal performance of our model always occurs in a wide range of \(\beta\) , i.e., [1e-6, 1e-3], [1e-3, 10], [1e-8, 1e-6], and [1e-6, 1e-3] for \textit{Indian Pines}, \textit{Salinas}, \textit{Pavia University}, and \textit{WHU-Hi-LongKou}, respectively. These results demonstrate the robustness of our method with respect to the parameter \(\beta\). Furthermore, it can be observed our model with nonzero $\beta$ always lead to a higher OA on all datasets which illustrates the effectiveness of the global regularization term.


\begin{table*}[ht]
\centering
\caption{Ablation Study on \textit{Salinas} with 1\% Training Samples. Optimal Values Are Denoted in Bold, and the Second-Best Values Are Underlined}
\scalebox{0.8}{
{\begin{tabular}{c|cc|ccccc|ccc|c} 
\hline\hline
\#     & Train&Test & Origin              & RPCA \citep{2011rpca}              &Super-RPCA& TRPCA \citep{2020TRPCA}             & Patch-TRPCA& M1                     & M2                       &M3& Proposed  
\\\hline
1 & 21&1988 & 97.55 ± 1.59& 97.10 ± 2.03 &98.91 ± 1.07& 96.90 ± 1.38 &95.42 ± 5.16 & 97.76 ± 2.05& \underline{99.42 ± 0.99} &\textbf{99.84 ± 0.07} & {97.78 ± 3.21}
\\
2 & 38&3688 & 99.30 ± 0.47& 99.04 ± 1.18 &99.49 ± 0.37& 96.68 ± 2.07 &98.62 ± 1.47 & 98.83 ± 0.49& \underline{99.64 ± 0.39} &\textbf{99.72 ± 0.18} & {99.62 ± 0.48}
\\
3 & 20&1956 & 96.61 ± 3.01& 95.96 ± 4.38 &95.81 ± 3.72& 96.96 ± 2.53 &98.30 ± 1.97 & 97.36 ± 2.18& {99.15 ± 1.11}  &\underline{99.44 ± 0.84} & \textbf{99.94 ± 0.12}
\\
4 & 14&1380 & \underline{99.22 ± 0.47} & \textbf{99.28 ± 0.37} &97.42 ± 1.88& 98.03 ± 2.19 &97.07 ± 3.28 & 96.91 ± 3.21& 93.68 ± 5.45 &96.39 ± 1.24 & 95.58 ± 2.90
\\
5 & 27&2651 & 96.94 ± 0.92& 96.83 ± 0.75 &96.88 ± 1.49& 96.46 ± 0.85 &95.35 ± 2.82 & {97.13 ± 1.26} & 96.78 ± 1.62 &\underline{97.37 ± 0.83} & \textbf{97.85 ± 0.95}
\\
6 & 40&3919 & 99.35 ± 0.73& 99.39 ± 0.53 &99.49 ± 0.28& 99.56 ± 0.28 &98.84 ± 0.86 & 99.79 ± 0.08& \underline{99.88 ± 0.03}  &99.06 ± 0.38 & \textbf{99.89 ± 0.11}
\\
7 & 36&3543 & {99.39 ± 0.16} & 99.27 ± 0.45 &99.42 ± 0.45& 94.88 ± 2.94 &97.39 ± 1.16 & 98.30 ± 1.07& 98.76 ± 0.71 &\textbf{99.58 ± 0.10} & \underline{99.51 ± 0.44}
\\
8 & 113& 11158 & 81.76 ± 2.93& 80.64 ± 1.99 &90.82 ± 1.88& 94.83 ± 2.29 &97.12 ± 0.99 & 94.83 ± 0.94& {95.97 ± 1.46}  &\textbf{97.90 ± 0.43} & \underline{97.74 ± 1.38}
\\
9 & 63&6140 & {99.24 ± 0.67} & 99.13 ± 0.81 &98.46 ± 0.84& 99.06 ± 0.59 &99.14 ± 0.82 & 98.35 ± 0.72& 98.83 ± 0.53 &\textbf{99.87 ± 0.12} & \underline{99.51 ± 0.38}
\\
10 & 33&3245 & 90.21 ± 1.99& 88.93 ± 3.25 &89.64 ± 3.15& {93.76 ± 2.26}  &89.59 ± 3.31 & 91.15 ± 2.14& 93.70 ± 2.83 &\underline{94.83 ± 1.89} & \textbf{95.39 ± 2.10}
\\
11 & 11&1057 & 90.46 ± 4.81& 91.09 ± 6.05 &91.71 ± 4.48& 93.43 ± 3.35 &88.68 ± 8.34 & 92.34 ± 4.52& \underline{96.38 ± 3.22}  &95.99 ± 1.23 & \textbf{96.67 ± 4.36}
\\
12 & 20&1907 & 99.65 ± 0.36& 99.49 ± 0.86 &99.66 ± 0.47& \underline{99.69 ± 0.15}  &96.20 ± 3.54 & \textbf{99.79 ± 0.38}& 99.58 ± 0.52 &99.47 ± 0.72 & 99.42 ± 0.79
\\
13 & 10&906 & {96.82 ± 2.55} & \textbf{97.25 ± 2.07} &\underline{96.89 ± 3.93}
& 96.51 ± 3.24 &91.39 ± 5.78 & 96.18 ± 1.82& 96.40 ± 1.82 &94.90 ± 2.36 & 95.91 ± 2.87
\\
14 & 11&1059 & 91.86 ± 3.50& {92.11 ± 2.81}  &89.80 ± 4.84& 91.35 ± 3.66 &77.58 ± 14.22 & 91.10 ± 7.06& 91.76 ± 7.45 &\textbf{96.45 ± 0.34} & \underline{96.39 ± 3.32}
\\
15 & 73&7195 & 67.61 ± 3.95& 68.74 ± 2.48 &79.94 ± 3.78& 94.23 ± 2.27 &96.53 ± 1.20 & 95.29 ± 1.64& \underline{96.07 ± 2.14}  &95.12 ± 1.78 & \textbf{97.44 ± 1.56}
\\
16 & 19&1788 & 96.12 ± 3.08& 94.99 ± 5.67 &94.97 ± 7.24& \underline{97.91 ± 0.83}  &98.19 ± 0.92 & 96.18 ± 3.06& 97.70 ± 2.34 &97.73 ± 0.66 & \textbf{98.09 ± 1.06}
\\\hline
\multicolumn{3}{c|}{OA} & 90.10 ± 0.51& 89.85 ± 0.46 &93.47 ± 0.59& 96.17 ± 0.67 &96.31 ± 0.53 & 96.43 ± 0.42& {97.24 ± 0.37}  &\underline{97.88 ± 0.19} & \textbf{98.19 ± 0.41}
\\
\multicolumn{3}{c|}{AA} & 93.88 ± 8.53& 93.70 ± 8.38 &94.96 ± 5.37& 96.26 ± 2.36 &94.71 ± 5.59 & 96.33 ± 2.78& {97.11 ± 2.47}  &\underline{97.73 ± 1.90} & \textbf{97.92 ± 1.60}
\\
\multicolumn{3}{c|}{$\kappa$} & 88.98 ± 0.56& 88.69 ± 0.51 &92.72 ± 0.66& 95.74 ± 0.74 &95.89 ± 0.59 & 96.03 ± 0.46& {96.93 ± 0.41}  &\underline{97.64 ± 0.22} & \textbf{97.98 ± 0.45}
\\
\hline\hline
\end{tabular}
}}
\label{study2}
\end{table*}

\begin{table*}[ht]
\centering
\caption{Ablation Study on \textit{Pavia University} with 0.5\% Training Samples. Optimal Values Are Denoted in Bold, and the Second-Best Values Are Underlined}
\scalebox{0.8}{
{\begin{tabular}{c|cc|ccccc|ccc|c} 
\hline\hline
\#     & Train&Test & Origin              & RPCA \citep{2011rpca}              &Super-RPCA& TRPCA \citep{2020TRPCA}             &Patch-TRPCA& M1                     & M2                       &M3& Proposed                      \\\hline
1 & 34&6597
& 85.66 ± 3.22& 83.61 ± 3.22 &84.92 ± 4.52& \textbf{88.92 ± 2.72} &85.04 ± 3.23 & 85.01 ± 3.99& 85.85 ± 3.25 &86.46 ± 2.86 & \underline{86.65 ± 2.90} 
\\
2 & 94&18555
& 93.90 ± 2.40& 93.81 ± 1.62 &93.82 ± 2.04& 95.95 ± 0.78 &97.57 ± 1.18& 97.27 ± 0.98& \underline{99.14 ± 0.68}  &95.96 ± 1.26& \textbf{99.25 ± 0.55}
\\
3 & 11&2088
& 63.30 ± 10.42& 61.40 ± 7.49 &55.04 ± 10.91& 67.16 ± 8.80 &65.09 ± 5.00& 70.88 ± 8.60& \underline{71.02 ± 9.38}  &70.77 ± 9.12
& \textbf{75.72 ± 7.40}
\\
4 & 16&3048
& 81.76 ± 5.69& 83.29 ± 4.80 &86.15 ± 2.45& 81.72 ± 5.39 &\underline{90.99 ± 3.18}& {90.78 ± 4.64} & 90.71 ± 3.90 &\textbf{93.56 ± 2.39}&{ 90.87 ± 2.64}
\\
5 & 7&1338
& 94.79 ± 10.24& 95.44 ± 7.59 &\textbf{99.21 ± 0.35}& {99.01 ± 0.46} &{99.12 ± 0.24}& 98.69 ± 0.93& {98.73 ± 0.70}  &\underline{99.18 ± 0.28}& 97.94 ± 4.70
\\
6 & 26&5003
& 74.12 ± 5.43& 72.14 ± 6.30 &71.04 ± 5.99& 79.99 ± 6.15 &80.36 ± 4.51& 80.20 ± 4.19& {91.69 ± 4.82}  &\underline{92.85 ± 2.73}& \textbf{95.26 ± 2.63}
\\
7 & 7&1323
& 68.70 ± 10.47& 73.17 ± 8.65 &73.94 ± 9.82& 79.92 ± 5.45 &75.90 ± 8.76& \underline{81.35 ± 8.88} & 80.42 ± 4.92 &75.33 ± 12.14
& \textbf{82.44 ± 8.04}
\\
8 & 19&3663
& 77.05 ± 6.35& 79.80 ± 5.84 &\underline{86.16 ± 2.59}
& 76.55 ± 8.10 &82.02 ± 4.67
& 84.76 ± 4.41& \underline{85.87 ± 6.27}  &78.98 ± 5.35
& \textbf{87.74 ± 3.87}
\\
9 & 5&942
& 99.72 ± 0.13& \textbf{99.76 ± 0.11} &\underline{99.81 ± 0.12}
& 99.05 ± 0.59 &99.66 ± 0.09
& 99.72 ± 0.09& \textbf{99.73 ± 0.11}  &99.68 ± 0.20
& 99.70 ± 0.27
\\\hline
\multicolumn{3}{c|}{OA}& 85.85 ± 1.41& 85.67 ± 0.91 &86.34 ± 1.09
& 88.55 ± 1.33 &89.62 ± 0.47 & 90.13 ± 1.02& \underline{92.50 ± 1.00} &90.79 ± 0.68 & \textbf{93.53 ± 0.73}
\\
\multicolumn{3}{c|}{AA} & 82.11 ± 12.48& 82.49 ± 12.46 &83.34 ± 14.54
& 85.36 ± 11.06 &86.19 ± 11.75 & 87.63 ± 9.76& \underline{89.24 ± 9.61} &88.08 ± 10.73 & \textbf{90.62 ± 8.26}
\\
\multicolumn{3}{c|}{$\kappa$} & 81.12 ± 1.83& 80.89 ± 1.23 &81.78 ± 1.41
& 84.71 ± 1.82 &86.14 ± 0.57 & 86.84 ± 1.37& \underline{90.01 ± 1.34} &87.84 ± 0.90 & \textbf{91.40 ± 0.97}
\\
\hline\hline
\end{tabular}
}}
\label{study3}
\end{table*}

\begin{table*}[ht]
\centering
\caption{Ablation Study on \textit{WHU-Hi-LongKou} with 1\% Training Samples. Optimal Values Are Denoted in Bold, and the Second-Best Values Are Underlined}
\scalebox{0.8}{
{\begin{tabular}{c|cc|ccccc|ccc|c} 
\hline\hline
\#     & Train&Test & Origin              & RPCA \citep{2011rpca}              &Super-RPCA& TRPCA \citep{2020TRPCA}              & Patch-TRPCA& M1                     & M2                       &M3& Proposed                      \\\hline
1 & 15&1396
& 88.38 ± 4.13& 87.97 ± 5.27 &86.92 ± 8.66
& 94.19 ± 2.46 &97.32 ± 2.81
& 99.27 ± 0.61
& \underline{98.91 ± 1.50}  &97.54 ± 3.21
& \textbf{99.08 ± 0.75}
\\
2 & 36&3532
& 87.88 ± 3.32& 86.98 ± 3.25 &88.23 ± 1.75
& 95.90 ± 1.85 &93.59 ± 3.18
& \textbf{98.79 ± 0.92}
& 97.80 ± 1.16 &95.14 ± 1.65
& \underline{98.09 ± 1.49} 
\\
3 & 13&1216
& 73.27 ± 6.44& 72.71 ± 5.25 &79.03 ± 3.38
& 87.36 ± 5.92 &93.93 ± 3.42
& \underline{95.61 ± 0.82}
& 92.03 ± 8.01 &\textbf{97.66 ± 0.32}
& {95.16 ± 1.75}
\\
4 & 59&5773
& 85.51 ± 2.83& 85.38 ± 2.54 &86.04 ± 1.56
& 91.44 ± 1.92 &94.32 ± 3.30
& 95.49 ± 1.43
& {95.84 ± 1.68}  &\textbf{97.75 ± 0.37}
& \underline{95.86 ± 1.79}
\\
5 & 14&1324
& 71.11 ± 5.00& 71.24 ± 7.80 &78.49 ± 3.52
& 74.90 ± 7.75 &86.42 ± 3.31
& \underline{94.89 ± 1.82}
& {93.15 ± 3.82}  &\textbf{97.39 ± 0.79}
& 92.64 ± 2.82
\\
6 & 7&665
& 87.43 ± 7.51& 84.98 ± 11.85 &86.59 ± 7.46
& 95.28 ± 5.79 &98.77 ± 2.03
& 99.91 ± 0.20
& \underline{99.92 ± 0.22}  &99.79 ± 0.33
& \textbf{99.98 ± 0.07}
\\
7 & 207&20403
& \underline{99.98 ± 0.01}& 99.98 ± 0.01 &\textbf{99.99 ± 0.00}
& 99.74 ± 0.30 &99.93 ± 0.04
& 99.87 ± 0.18
& {99.95 ± 0.03}  &99.94 ± 0.01
& 99.94 ± 0.04
\\
8 & 25&2428
& 79.61 ± 6.65& 81.45 ± 4.07 &81.87 ± 7.13
& 84.62 ± 5.16 &84.67 ± 2.36
& 86.47 ± 5.48
& \textbf{90.40 ± 4.61} &84.22 ± 6.80
& \underline{90.23 ± 3.39} 
\\
9 & 15&1435
& 55.93 ± 5.65& 59.87 ± 6.52 &55.89 ± 3.98
& 63.21 ± 8.09 &\textbf{83.47 ± 3.64}
& 67.33 ± 10.34
& 75.52 ± 6.56 &73.90 ± 4.31
& \underline{80.33 ± 8.28}
\\\hline
\multicolumn{3}{c|}{OA}& 91.23 ± 0.66& 91.32 ± 0.79 &91.86 ± 0.47
& 94.26 ± 0.60 &96.13 ± 0.84 & 96.70 ± 0.65 & \underline{97.08 ± 0.55}  &96.93 ± 0.37 & \textbf{97.36 ± 0.46}
\\
\multicolumn{3}{c|}{AA}& 81.01 ± 12.81& 81.18 ± 11.67 &82.56 ± 11.85
& 87.40 ± 11.72 &92.49 ± 6.17 & 93.07 ± 10.52 & \underline{93.72 ± 7.67} &93.70 ± 8.81 & \textbf{94.59 ± 6.31}
\\
\multicolumn{3}{c|}{$\kappa$}& 86.96 ± 0.97& 87.09 ± 1.18 &87.90 ± 0.70
& 91.47 ± 0.89 &94.25 ± 1.23 & 95.10 ± 0.96 & \underline{95.65 ± 0.81}  &95.44 ± 0.56 &\textbf{ 96.08 ± 0.68}
\\
\hline\hline
\end{tabular}}
}
\label{study4}
\end{table*}

\subsection{Ablation Study}

An ablation study was conducted to evaluate the effectiveness of different modules in the proposed method. First, we define the ITLRR model (Eq. (\ref{Eq1})), which employs the tensor nuclear norm as outlined in Definition \ref{definition_nuclear_norm}, and refer to it as M1. The model described in Eq. (\ref{EQ_combined}), which utilizes a non-convex tensor norm, is designated as M2. M3 is the extension of M1 with the addition of the global regularization term. {Eight models, including RPCA \citep{2011rpca}, Super-RPCA, TRPCA \citep{2020TRPCA}, Patch-TRPCA, M1, M2, M3 and the final proposed model (Eq. (\ref{EQ3})), are tested. The Super-RPCA method applies superpixel segmentation to extract homogeneous regions, which are then unfolded into matrices and processed using RPCA \citep{2011rpca}. Similarly, the Patch-TRPCA method divides the data into multiple regular 3D tensor blocks, each independently processed using TRPCA \citep{2020TRPCA}.} The classification accuracy of the original data is also listed as a reference.

{As listed in Tables \ref{study1}-\ref{study4}, there is only a slight improvement in RPCA compared to the original data. By capturing local information, Super-RPCA achieves substantial performance improvement on \textit{Indian Pines} and \textit{Salinas} compared to RPCA. In contrast, TRPCA significantly improves performance by preventing information loss typically caused by unfolding 3D data. Patch-TRPCA further enhances performance on \textit{Salinas}, \textit{Pavia University}, and \textit{WHU-Hi-LongKou}, as it can focus on more localized features. However, due to its simplistic block division approach, a performance decline is observed on \textit{Indian Pines}, which has a more complex shape. }

In contrast, our proposed ITLRR model demonstrates performance enhancements across all four datasets compared to the TRPCA model and also outperforms the Patch-TRPCA model in each dataset by utilizing local information better. Notably, when compared to TRPCA, M1 demonstrates a more substantial improvement in AA. This improvement is particularly pronounced for categories with fewer pixels, such as classes 1, 7, 9, and 16 on \textit{Indian Pines}, highlighting the effectiveness of the irregular low-rank tensor representation introduced by M1.

Compared with M1, M2 further improves performance by utilizing a non-convex tensor norm that approximates rank function better. It is evident that the classification performance of \textit{Salinas} and \textit{Pavia University} has been significantly improved, which showed less improvement using M1. Especially, in comparison to the OA values of TRPCA on \textit{Salinas} (96.17\%) and \textit{Pavia University} (88.55\%), M2 achieves higher accuracy with values of 97.24\% and 92.50\%, while the ones of M1 are only 96.43\% and 90.13\%. 

{To mitigate the issue of local over-smoothing, which arises from solely focusing on local low-rank constraints, we introduce a global regularization term to enhance the discriminability between classes. It can be observed that M3, which incorporates the global regularization term based on M1, achieves improved performance across four datasets. By adding the global regularization term, M3 improves the accuracy significantly for classes with limited training samples, i.e., classes 1, 4, 7, and 15 on \textit{Indian Pines} and classes 1, 10, 11, and 14 on \textit{Salinas}. Specifically, on \textit{Indian Pines}, class 7 has only three training samples, yet the accuracy increases from 91.80\% to 96.80\%. Similarly, on \textit{Salinas}, the accuracy of class 14, which has only 11 training samples, rises from 91.10\% to 96.45\%.}

Finally, the final model with the non-convex tensor norm and the global regularization term achieves the highest values across all metrics on all four datasets. Especially, on \textit{Indian Pines}, the final model achieves the optimal accuracy in 12 out of 16 classes.


\begin{figure}[ht]
\centering
\subfigure[]{\includegraphics[width=0.24\textwidth]{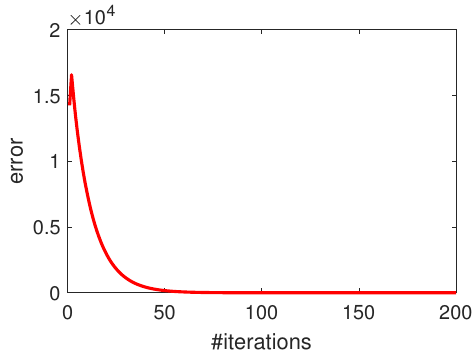}\label{conver1}}%
\hfil
\subfigure[]{\includegraphics[width=0.24\textwidth]{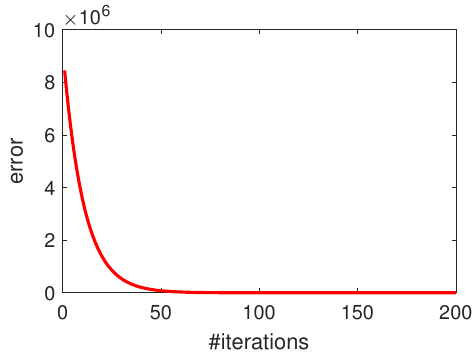}\label{conver2}}%

\subfigure[]{\includegraphics[width=0.24\textwidth]{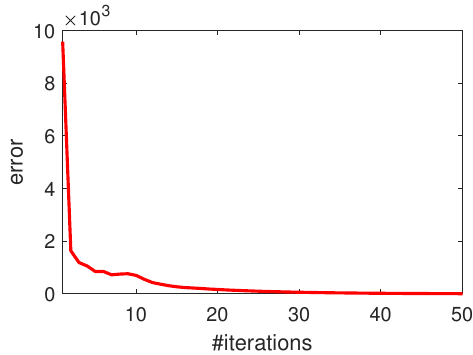}\label{conver3}}%
\hfil
\subfigure[]{\includegraphics[width=0.24\textwidth]{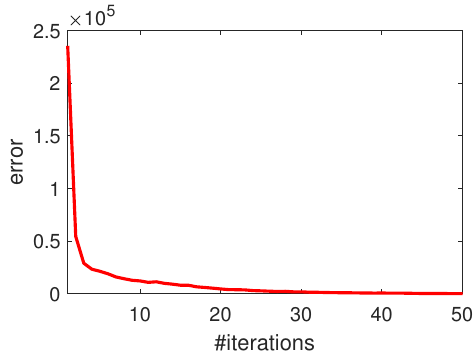}\label{conver4}}%
\caption{Convergence curves of proposed model on four datasets. (a) \textit{Indian Pines}. (b) \textit{Salinas}. (c) \textit{Pavia University}. (d) \textit{WHU-Hi-LongKou}.}
\label{fig_convergence}
\end{figure}

\subsection{Convergence Analysis} 

Fig. \ref{fig_convergence} shows the convergence curves of our model on all four datasets. For each subfigure, the Y-axis represents the convergence metric, defined as $\max(\|\mathcal{L}^{o^{t+1}} - \mathcal{L}^{o^{t}}\|_\infty, \|\mathcal{S}^{o^{t+1}} - \mathcal{S}^{o^{t}}\|_\infty, \|\mathcal{X} - \mathcal{L}^{o^{t+1}} - \mathcal{S}^{o^{t+1}}\|_\infty)$, which is used to check the convergence condition. We can see that the convergence metric decreases consistently with iteration goes on and finally approaches zero on all four datasets, which validates the empirical convergence of our proposed method.

\section{Conclusion}\label{sec:conclusion}

In this paper, we have presented a novel irregular tensor low-rank representation model. In contrast to existing models, our model is the first one to pursue discriminative low-rank representation for irregular data cubes without unfolding three-dimensional tensors into two-dimensional matrices. This allows our model to effectively capture the local spatial information of the tensors. Furthermore, we incorporate a global regularization term to enhance the discriminative ability of the representation. Lastly, we provide an iterative algorithm to efficiently solve the proposed problem with excellent empirical convergence. Experimental results on four widely-used datasets show that our model significantly outperforms state-of-the-art methods. In future work, we plan to integrate deep learning techniques to further enhance performance and explore broader applications of irregular tensor representations in downstream tasks.

\footnotesize
\bibliographystyle{IEEEtranN}
\bibliography{reference.bib}

\begin{IEEEbiography}[{\includegraphics[width=1.2in,height=1.25in,keepaspectratio]{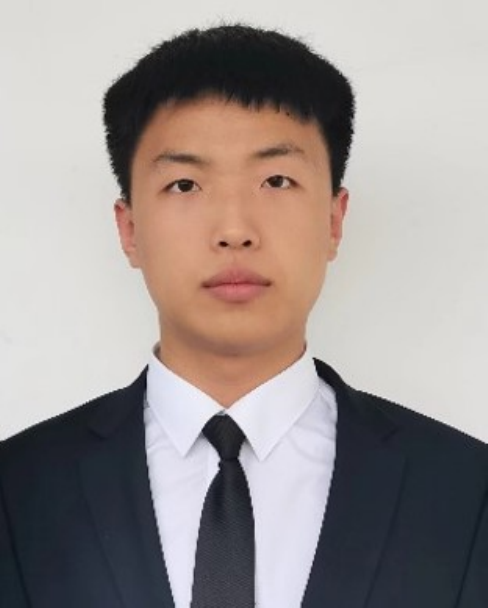}}]{Bo Han} received the B.S. degree in computer science and technology from Southeast University, Nanjing, China, in 2024. He is currently pursuing the M.S. degree in intelligent science and technology with Southeast University, Nanjing, China. His research interests include weakly supervised learning and multi-label learning.
\end{IEEEbiography}

\begin{IEEEbiography}[{\includegraphics[width=1in,height=1.25in,clip,keepaspectratio]{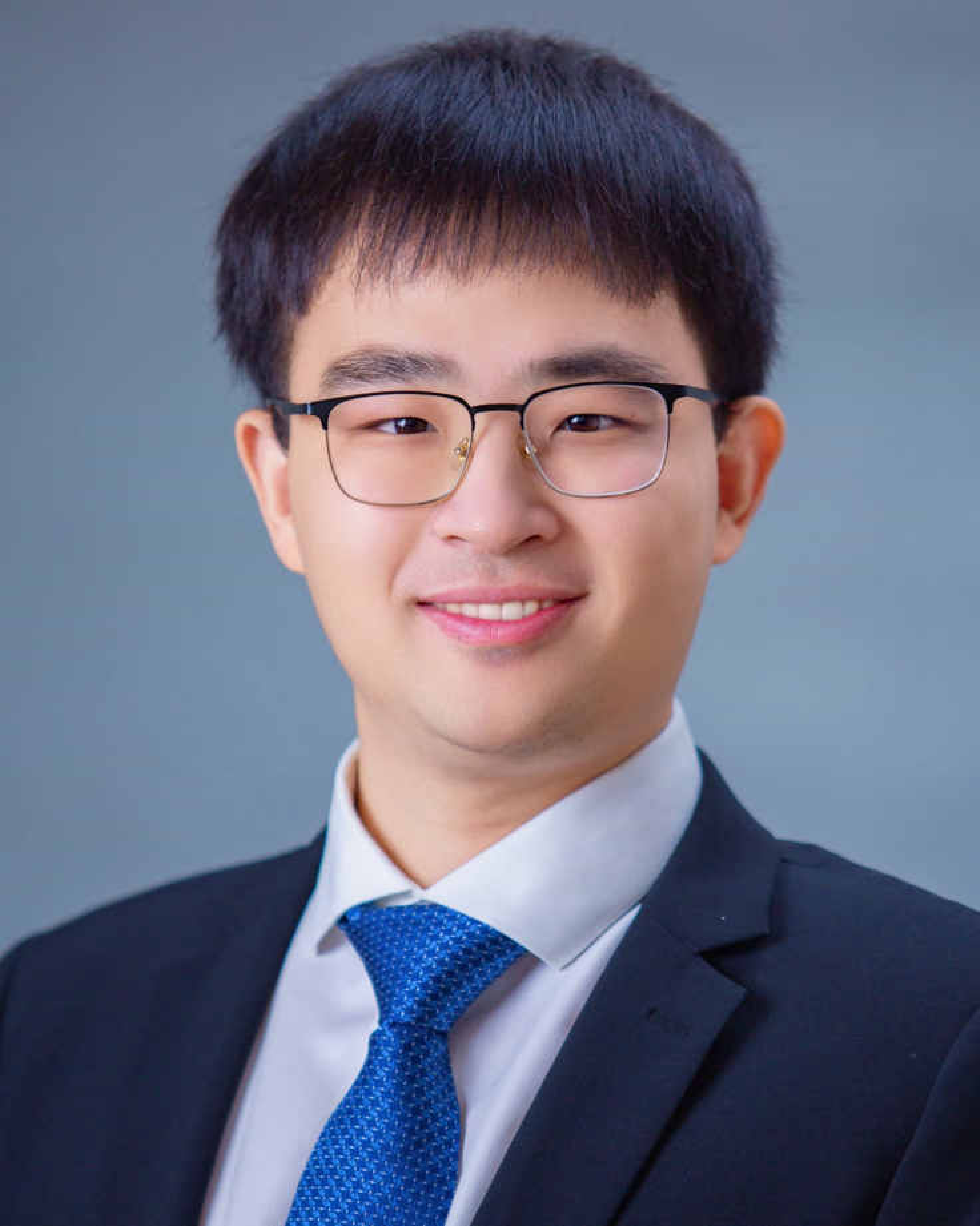}}]{Yuheng Jia} (Member, IEEE) received the B.S. degree in automation and the M.S. degree in control theory and engineering from Zhengzhou University, Zhengzhou, China, in 2012 and 2015, respectively, and the Ph.D. degree in computer science from the City University of Hong Kong, Hong Kong, China, in 2019.

He is currently an Associate Professor with the School of Computer Science and Engineering, Southeast University, Nanjing, China. His current research interests include machine learning and data representation, such as weakly-supervised learning, high-dimensional data modeling and analysis, and low-rank tensor/matrix approximation and factorization.
\end{IEEEbiography}

\begin{IEEEbiography}[{\includegraphics[width=1in,height=1.25in,clip,keepaspectratio]{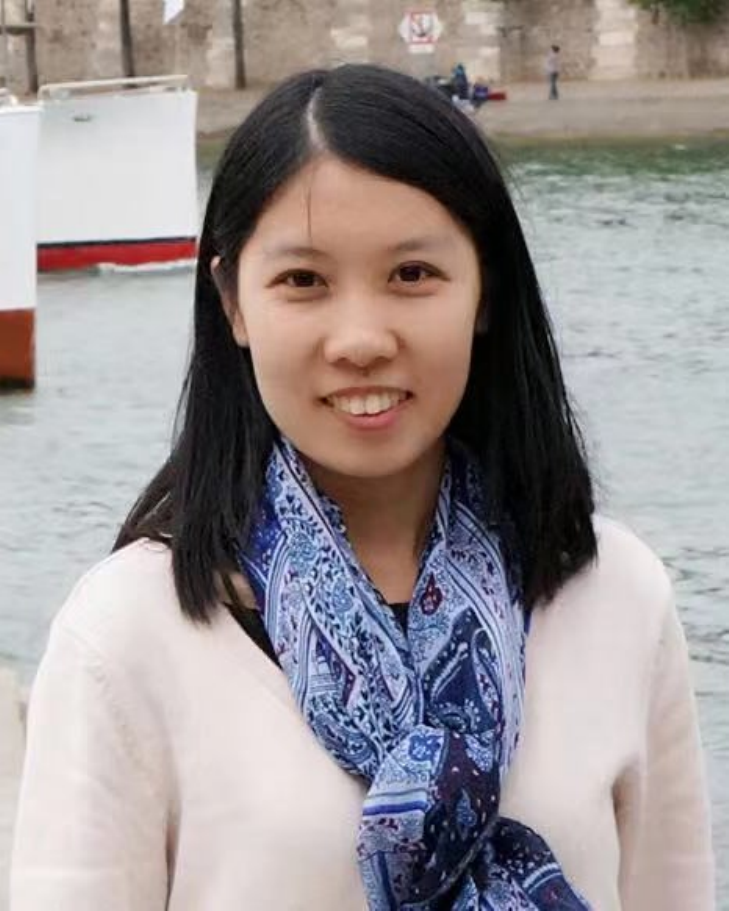}}]{Hui Liu} received the B.Sc. degree in communication engineering from Central South University, Changsha, China, the M.Eng. degree in computer science from Nanyang Technological University, Singapore, and the Ph.D. degree in computer science from the City University of Hong Kong, Hong Kong. From 2014 to 2017, she was a Research Associate with the Maritime Institute, Nanyang Technological University. She is currently an Assistant Professor with the Department of Computing and Information Sciences, Saint Francis University, Hong Kong. Her current research interests include image processing and machine learning.
\end{IEEEbiography}

\begin{IEEEbiography}[{\includegraphics[width=1in,height=1.25in,clip,keepaspectratio]{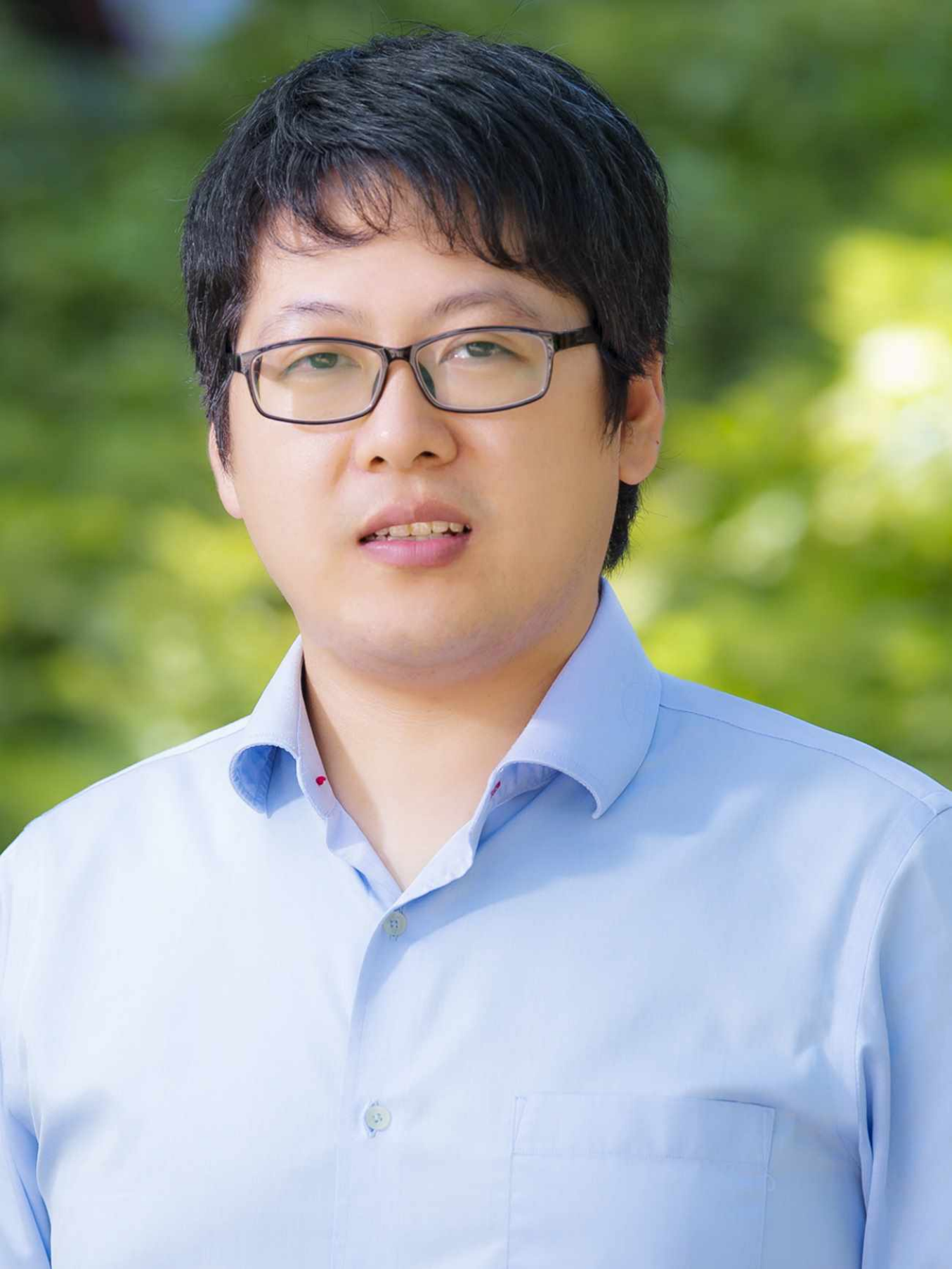}}]{Junhui Hou} (Senior Member, IEEE) is an Associate Professor with the Department of Computer Science, City University of Hong Kong. He holds a B.Eng. degree in information engineering (Talented Students Program) from the South China University of Technology, Guangzhou, China (2009), an M.Eng. degree in signal and information processing from Northwestern Polytechnical University, Xi’an, China (2012), and a Ph.D. degree from the School of Electrical and Electronic Engineering, Nanyang Technological University, Singapore (2016). His research interests are multi-dimensional visual computing.

Dr. Hou received the Early Career Award from the Hong Kong Research Grants Council in 2018 and the NSFC Excellent Young Scientists Fund in 2024. He has served or is serving as an Associate Editor for \textit{IEEE Transactions on Visualization and Computer Graphics}, \textit{IEEE Transactions on Image Processing}, \textit{IEEE Transactions on Multimedia}, and \textit{IEEE Transactions on Circuits and Systems for Video Technology}. 
\end{IEEEbiography}

\vfill

\end{document}